\PassOptionsToPackage{svgnames}{xcolor}
\documentclass[12pt]{article}

\usepackage{fontspec}

\usepackage{amsmath,amssymb,amsfonts,amsthm,setspace,tikz,natbib,dsfont,wrapfig,lipsum, xcolor, array,colortbl,mathrsfs,pifont,wasysym}
\usetikzlibrary{calc}
\usepackage{tikz-cd} 
\usetikzlibrary{patterns,decorations.pathreplacing,patterns.meta}

\usepackage{centernot}
\usepackage{xspace}
\usepackage{booktabs}
\usepackage{bm}
\usepackage{quoting}
\usepackage{subcaption}
\usepackage{thmtools,thm-restate}
\usepackage{enumitem}

\usepackage[makeroom]{cancel}

\usepackage[a4paper]{geometry}
\definecolor{lam1}{HTML}{E62758}
\definecolor{lam2}{HTML}{176C89}
\definecolor{lam3}{HTML}{2F701E}
\definecolor{lam4}{HTML}{591756}

\usepackage[colorlinks=true,urlcolor=lam2,linkcolor=lam3,citecolor=lam4]{hyperref}

\newtheoremstyle{mytheoremstyle} 
    {.3em}                    
    {.3em}                    
    {\itshape}                   
    {}                           
    {\sc}                   
    {.}                          
    {.3em}                       
    {}  

\theoremstyle{mytheoremstyle}

\newtheorem{definition}{Definition}

\newtheorem{theorem}{Theorem}
\newtheorem{lemma}[theorem]{Lemma}

\newtheorem{corollary}[theorem]{Corollary}

\newtheoremstyle{scfont} 
    {.3em}                    
    {.3em}                    
    {}                   
    {}                           
    {\scshape}                   
    {}                          
    {.3em}                       
    {\textbf{A\thmnumber{#2}\thmname{#1}}\thmnote{---{\textsc{#3}.}}}

\theoremstyle{scfont}

\theoremstyle{remark}   

\newtheorem{example}{Example}
\newtheorem*{example*}{Example}

\newenvironment{tproof}[1]{%
  \begin{proof}[Proof of Theorem \ref{#1}]%
  \phantomsection%
  \addcontentsline{toc}{subsection}{Proof of Theorem \ref{#1}}%
  \label{pf:#1}%
}{%
  \end{proof}%
}

\newenvironment{subproof}[1][\proofname]{
  \begin{proof}[#1]%
}{%
  \end{proof}%
}

\makeatletter
\newcommand{\mylabel}[3]{\def\@currentlabel{#2}\phantomsection\textsc{#3} (\texttt{#2})\label{#1}}
\newcommand{\sclabel}[2]{\def\@currentlabel{\theaxiom}\phantomsection\lowercase{\textsc{#2}}\label{ax:#1}}
\makeatother

\newcounter{axiom}
\newenvironment{axiom}[3]{
\smallskip
\addtocounter{axiom}{1}
\noindent (\textbf{#2}) -- \sclabel{#2}{#3}.
}{
\smallskip
}

\newcounter{thmp}

\let\OLDthebibliography\thebibliography
\renewcommand\thebibliography[1]{
  \OLDthebibliography{#1}
  \setlength{\parskip}{0pt}
  \setlength{\itemsep}{1pt plus 0.3ex}
}

\usepackage{titlesec}

\titleformat*{\section}{\large\bfseries}
\titleformat{\subsection}[runin]
  {\bfseries}           
  {\thesubsection}      
  {0.75em}              
  {}[. \ \  ]                 

\titlespacing*{\section}{0pt}{1.5ex plus 1ex minus .2ex}{1ex plus .2ex}
\titlespacing*{\subsection}{0pt}{1ex plus 1ex minus .2ex}{0ex plus .2ex}

\renewcommand{\emptyset}{\varnothing}
\renewcommand{\phi}{\varphi}

\newcommand{\commentout}[1]{}

\renewcommand{\implies}{\, {\Rightarrow}\, }
\renewcommand{\iff}{\, {\Leftrightarrow}\, }

\def \W{W}
\def \w{w}
\def \l{\mu}

\def \D{\Delta}
\def \R{\mathbb{R}}

\def \p{\piv}
\def \q{\textsc{q}}

\def \ff{\textsc{f}}

\def \f{\bm{f}}
\def \g{\bm{g}}

\def \L{\mathcal{L}}
\def \K{\mathcal{K}}

\def \s{\succcurlyeq}
\def\ss{\s^{\!\circ}}
\def\impp{\imp^{\!\circ}}
\def\notimpp{{\centernot{\imp}}^{\!\circ}}

\def \1{\mathds{1}}
\newcommand{\df}[1]{\textup{\textbf{#1}}}

\def\red{\color{lam1}\textsc{r}\color{black}\xspace}
\def\blue{\color{lam2}\textsc{b}\color{black}\xspace}

\def\actF{\color{lam1}f\color{black}\xspace}
\def\actG{\color{lam2}g\color{black}\xspace}

\def\actThree{\color{lam2}f_{3}\color{black}\xspace}
\def\actFour{\color{lam1}f_{4}\color{black}\xspace}

\def\state{\textsc{s}}
\def\vote{\textsc{v}}
\def\obs{\textsc{o}}
\def\piv{\textsc{p}}

\def\tt{\textsc{t}}

\def\aa{\textsc{a}}
\def\bb{\textsc{b}}
\def\cc{\textsc{c}}
\def\dd{\textsc{d}}

\setlist[enumerate]{itemsep=0pt, topsep=5pt}
\setlist[itemize]{itemsep=0pt, topsep=5pt}
\quotingsetup{vskip=3pt}

\makeatletter
\newcommand{\shorteq}{%
  \settowidth{\@tempdima}{..}
  \resizebox{\@tempdima}{\height}{$=$}%
}
\makeatother
\DeclareMathOperator \imp{\shorteq\!\!\s}

\usepackage{accents}

\newcounter{inline}

\newcommand{\refax}[1]{\textbf{\hyperref[ax:#1]{\textup{\uppercase{#1}}}}}
\newcommand{\refa}[1]{\textbf{\hyperref[ax:#1]{\textup{A\ref{ax:#1}}}}}
\newcommand{\refr}[1]{\textbf{\hyperref[ax:#1]{\textup{R\ref{ax:#1}}}}}

\newcommand{\cons}[1]{\{\!\!\{#1\}\!\!\}}
\newcommand{\blank}[1]{}

\newcommand{\dm}{DM\xspace}
\newcommand{\dms}{DMs\xspace}

\newcommand{\hide}[1]
{
}

\newcommand{\markcells}[4][]{%
  \foreach \st/\row in {#2}{%
    \fill[#1, draw=none, fill=#3]              
         ({\st-1},{\row-1}) rectangle ++(1,1);
    \node at ({\st-0.5},{\row-0.5}) {#4};   
  }%
}

\usepackage{silence}
\title{\vspace{-2.5cm}
Failures of Contingent Thinking\thanks{
  The authors thank Pierpaolo Battigalli, Emiliano Catonini, Christoph Kuzmics, Joe Halpern, Muriel Niederle, and Emanuel Vespa, as well as audiences at various conferences and seminars.
  Peio Zuazo-Garin acknowledges financial support from the Russian Academic Excellence Project “5-100.”
}}

\author{
Evan Piermont\footnote{Royal Holloway--University of London, United Kingdom; \href{mailto:evan.piermont@rhul.ac.uk}{\texttt{ evan.piermont@rhul.ac.uk}}.}
\hspace{1cm}
Peio Zuazo-Garin\footnote{NYU Shanghai, People's Republic of China; \href{mailto:peio.zuazo.garin@nyu.edu}{\texttt{peio.zuazo.garin@nyu.edu}}.}} 

\begin{document}

\setlength{\abovedisplayskip}{5pt}
\setlength{\belowdisplayskip}{5pt}
\setlength{\itemsep}{1pt plus 0.3ex}

\maketitle

\vspace{-1cm}

\begin{abstract}
\footnotesize
We present a behavioral definition of an agent's perceived implication that uniquely identifies a subjective state-space representing her view of a decision problem, and which may differ from the modeler's. By examining belief updating within this model, we formalize the recent empirical consensus that reducing uncertainty improves contingent thinking, and propose a novel form of updating corresponding to the agent `realizing' a flaw in her own thinking.
 Finally, we clarify the sense in which contingent thinking makes state-by-state dominance more cognitively demanding than obvious dominance.


\end{abstract}


\section{Introduction}
\label{section:motivation}

There are many diverse ways that economic agents misinterpret or misperceive the true decision problem they face,\footnote{E.g., assessing likelihoods in a manner incompatible with probability \citep{kahneman1979prospect,schmeidler-89,tversky1994support,walkerjones2023nonBayesian}, disregarding the information-generating process \citep{jin2015no, enke2019correlation, enke2020you}, failing to properly condition beliefs \citep{tversky1983extensional, thaler1988anomalies, esponda2014hypothetical}, ignoring other agents' strategic considerations \citep{eyster2005cursed, esponda2008behavioral}, etc.} 
and many of these empirical phenomena have been directly attributed to flawed contingent thinking.%
\footnote{E.g., 
\cite{esponda2024contingent}: ``subjects are not good at thinking through the state-space in the way analysts often assume and [...] incomplete preferences or anomalies may precisely stem from the fact that states are not naturally given, may be hard to construct, or some states may not be salient;'' 
\cite{charness2009origin}: ``the origin of the winner's curse does not lie in current theoretical explanations [but from subjects' failure] to recognize that a future contingency is relevant for their current decisions;'' 
\cite{martinez2019failures}: ``aggregating over multiple possible values of the state is especially difficult when there is uncertainty;" 
\cite{calford2022contingent}: ``many [subjects] have difficulty in effectively solving contingent thinking problems optimally.''} 
Nonetheless, there exists no universal definition of `contingent thinking' nor a unifying framework to collectively analyze these examples. This paper offers such a framework.
We posit a simple definition of \emph{subjective implication} to analyze a decision maker (\dm) whose understanding of a decision problem may diverge from that of the modeler or experimenter, and which may be logically
unsound. Our main result uses this definition to identify, from observable betting behavior, a wholly subjective state-space
capturing the DM’s (possibly unsound) view of the decision problem.
We then apply our methodology to study the dynamics of contingent thinking and how flawed reasoning can give rise to violations of state-wise dominance.

The primitive of our theory is a set of linguistic statements such as ``the value of the stock exceeds \$10'' or ``the ball drawn from the urn is red,'' and the observable data is the \dm's preferences regarding \emph{bets}---contingent claims that yield a specified payoff if any statement in a fixed set is true, and nothing otherwise. From this preference, we advance our central conceptual novelty: a definition of \emph{subjective implication}.

 A \dm \emph{perceives that $\phi$ implies $\psi$} if for any set of winning conditions $\Psi$ containing $\psi$, she is indifferent between betting on $\Psi$ or $\Psi \cup \{\phi\}$. 
 Intuitively, if the \dm perceives $\phi$ to imply $\psi$, she believes that whenever $\phi$ is true, so too is $\psi$; accordingly, if she already receives a payoff contingent on $\psi$, she will see no additional value in the winning criterion $\phi$.

Our main results (Theorems \ref{thm:iounoprob} and \ref{thm:maxstatespace}) establish that 
the \dm's subjective implication identifies her \emph{interpretation of uncertainty} (IOU): a pair $(\W,t)$ consisting of a \emph{state-space} $\W$, and a \emph{truth assignment} $t$ that maps each statement $\phi$ to the set of states, $t(\phi) \subseteq \W$, where $\phi$ is interpreted to be true. An IOU must faithfully recount the \dm's understanding of implication: $t(\phi)\subseteq t(\psi)$ exactly when the \dm perceives $\phi$ to imply $\psi$.
So, from the \dm's preferences over bets, we procure a comprehensive view into the \dm's understanding of how uncertainty might resolve.
For example, if $t(\phi)\cap t(\psi) = \emptyset$, then the \dm believes that $\phi$ and $\psi$ are incompatible; if $t(\phi)\cup t(\phi') = t(\psi)$, then she believes that $\phi$ and $\phi'$ are mutually equivalent to $\psi$, etc. 
As shown in the following example, this allows for distinguishing between different explanations of observed behavior.

\begin{example}
\label{ex:kt}
\cite{tversky1983extensional} provided the following vignette: 
\begin{quoting}
\small
Linda is 31 years old, single, outspoken, and very bright. She majored in philosophy. As a student, she was deeply concerned with issues of discrimination and social justice, and also participated in anti-nuclear demonstrations.
\end{quoting}
and asked subjects to rank the following statements in order of likelihood:
\vspace{-1ex}
\begin{itemize}[leftmargin=2cm, labelsep=.2em, itemsep=-1.4ex]
\small
\item[$\ff  =$]``Linda is active in the feminist movement.''
\item[$\tt =$]``Linda is a bank teller.''
\item[$\tt \land \ff  =$]``Linda is a bank teller and is active in the feminist movement.''
\end{itemize}

Eighty-five percent of subjects ranked $\ff$ more likely than $\tt \land\ff $ more likely than $\tt$. It then stands to reason that subjects would \emph{strictly} prefer to bet on $\{\tt, \tt \land \ff\}$ rather than $\{\tt\}$, indicating that they do not perceive that $\tt \land \ff$ implies $\tt$. 
Such betting behavior would reveal that for the subjects' interpretation of uncertainty, $t(\tt \land \ff) \centernot{\subseteq} t(\tt)$;
in violation of elementary logic, they consider a state where $\tt \land \ff$ is true but $\tt$ is not. 
Consequently, whereas \citeauthor{tversky1983extensional}'s experimental design reveals this misperception only indirectly, our framework provides a direct test via subjective implication. As argued next, it also allows for discriminating among competing theories of flawed reasoning.

\begin{figure}[]
  \centering
  \begin{subfigure}[t]{0.3\textwidth}{}
    \centering
    \begin{tikzpicture}[
                every node/.style   ={font=\footnotesize},
        gridlines/.style    ={gray,dotted,thin},
        x=.9cm,
        y=0.4cm
    ]
      \def\n{4}
      \def\m{3}

    \definecolor{tt}{HTML}{c45b5b}
      \markcells{1/1, 2/1}{tt!50}{$\tt$}
      \markcells[pattern={Lines[angle=45,distance={1.2pt},line width={0.6pt}]}, pattern color=tt!30]{3/1, 4/1}{}{}

      \definecolor{ff}{HTML}{5186b5}
      \markcells{3/2, 2/2}{ff!50}{$\ff$}
      \markcells[pattern={Lines[angle=45,distance={1.2pt},line width={0.6pt}]}, pattern color=ff!30]{1/2, 4/2}{}{}

      \definecolor{tf}{HTML}{d4ad00}
      \markcells{2/3}{tf!50}{$\tt {\land} \ff$}
      \markcells[pattern={Lines[angle=45,distance={1.2pt},line width={0.6pt}]}, pattern color=tf!30]{1/3, 3/3, 4/3}{}{}

      \foreach \x in {-1,...,\numexpr\n-1}{
          \draw[gridlines] (\x+1,-1) -- (\x+1,\m+.5);
      }
      \foreach \y in {-1,...,\numexpr\m-1}{
          \draw[gridlines] (-.3,\y+1) -- (\n+.3,\y+1);
      }

      \draw[thick] (0,0) rectangle (\n,\m);
      \foreach \i in {1,...,\numexpr\n-1}{
          \draw (\i,0) -- (\i,\m);
      }

      \draw[very thick] (-0.25,0) -- (\n+0.25,0);

      \foreach \i in {0,...,\numexpr\n-1}{
           \node[above=4pt] at (\i+0.5,-1.7) {$w_{\number\numexpr\i+1}$};
      }
    \end{tikzpicture}
    \caption{IOU $(W, t^\circ)$}
    \label{fig:state-space-Linda-Rat}
  \end{subfigure}
  \hfill
  \begin{subfigure}[t]{0.3\textwidth}{}
    \centering
    \begin{tikzpicture}[
                every node/.style   ={font=\footnotesize},
        gridlines/.style    ={gray,dotted,thin},
        x=.9cm,
        y=0.4cm
    ]
      \def\n{4}
      \def\m{3}

    \definecolor{tt}{HTML}{c45b5b}
      \markcells{1/1}{tt!50}{$\tt$}
      \markcells[pattern={Lines[angle=45,distance={1.2pt},line width={0.6pt}]}, pattern color=tt!30]{2/1, 3/1, 4/1}{}{}

      \definecolor{ff}{HTML}{5186b5}
      \markcells{3/2}{ff!50}{$\ff$}
      \markcells[pattern={Lines[angle=45,distance={1.2pt},line width={0.6pt}]}, pattern color=ff!30]{1/2, 2/2, 4/2}{}{}

      \definecolor{tf}{HTML}{d4ad00}
      \markcells{2/3}{tf!50}{$\tt {\land} \ff$}
      \markcells[pattern={Lines[angle=45,distance={1.2pt},line width={0.6pt}]}, pattern color=tf!30]{1/3, 3/3, 4/3}{}{}

      \foreach \x in {-1,...,\numexpr\n-1}{
          \draw[gridlines] (\x+1,-1) -- (\x+1,\m+.5);
      }
      \foreach \y in {-1,...,\numexpr\m-1}{
          \draw[gridlines] (-.3,\y+1) -- (\n+.3,\y+1);
      }

      \draw[thick] (0,0) rectangle (\n,\m);
      \foreach \i in {1,...,\numexpr\n-1}{
          \draw (\i,0) -- (\i,\m);
      }

      \draw[very thick] (-0.25,0) -- (\n+0.25,0);

      \foreach \i in {0,...,\numexpr\n-1}{
           \node[above=4pt] at (\i+0.5,-1.7) {$w_{\number\numexpr\i+1}$};
      }
    \end{tikzpicture}
    \caption{IOU $(\W,t)$}
    \label{fig:state-space-Linda-A}
  \end{subfigure}
  \hfill
  \begin{subfigure}[t]{0.3\textwidth}
    \centering
    \begin{tikzpicture}[
                every node/.style   ={font=\footnotesize},
        gridlines/.style    ={gray,dotted,thin},
        x=.9cm,
        y=0.4cm
    ]
      \def\n{3}
      \def\m{3}

      \definecolor{tt}{HTML}{c45b5b}
      \markcells{1/1}{tt!50}{$\tt$}
      \markcells[pattern={Lines[angle=45,distance={1.2pt},line width={0.6pt}]}, pattern color=tt!30]{2/1, 3/1}{}{}

      \definecolor{ff}{HTML}{5186b5}
      \markcells{2/2}{ff!50}{$\ff$}
      \markcells[pattern={Lines[angle=45,distance={1.2pt},line width={0.6pt}]}, pattern color=ff!30]{1/2, 3/2}{}{}

      \definecolor{tf}{HTML}{d4ad00}
      \markcells{1/3,2/3}{tf!50}{$\tt {\land} \ff$}
      \markcells[pattern={Lines[angle=45,distance={1.2pt},line width={0.6pt}]}, pattern color=tf!30]{3/3}{}{}

      \foreach \x in {-1,...,\numexpr\n-1}{
          \draw[gridlines] (\x+1,-1) -- (\x+1,\m+.5);
      }
      \foreach \y in {-1,...,\numexpr\m-1}{
          \draw[gridlines] (-.3,\y+1) -- (\n+.3,\y+1);
      }

      \draw[thick] (0,0) rectangle (\n,\m);
      \foreach \i in {1,...,\numexpr\n-1}{
          \draw (\i,0) -- (\i,\m);
      }

      \draw[very thick] (-0.25,0) -- (\n+0.25,0);

      \foreach \i in {0,...,\numexpr\n-1}{
          \node[above=4pt] at (\i+0.5,-1.7) {$w'_{\number\numexpr\i+1}$};
      }
    \end{tikzpicture}
    \caption{IOU $(\W',t')$}
    \label{fig:state-space-Linda-B}
  \end{subfigure}

  \caption{Three IOUs representing the statements from \cite{tversky1983extensional}. Each statement corresponds to a row; the truth assignment maps each statement to the states in its row with filled in cells. The first panel is the `objective' state-space and the latter two are logically flawed but consistent with the experimental findings.}
  \label{fig:state-space-pair}
\end{figure}


One proposed explanation for the observed response is that subjects perceive the statement ``Linda is a bank teller'' as ``Linda is a bank teller \emph{and not active in the feminist movement}'' and make the analogous inference for $\ff$.\footnote{Such implied, but unspoken, qualifications of meaning---referred to by linguists as \emph{conversational implicature}---are colloquially ubiquitous: ``These are my friends Ani and John, John is from Australia'' is widely understood to imply Ani is \emph{not} from Australia.  \cite{dulany1991conversational} and \cite{hilton1995social} relate this to the conjunction fallacy.} This  can be represented by the subjective interpretation of uncertainty $(\W,t)$ with $\W = \{\w_1, \w_2, \w_3,\w_4\}$ and $t: \tt  \mapsto \{\w_1\}, \tt\land\ff \mapsto \{\w_2\}$ and $\ff \mapsto \{\w_3\}$, shown in Figure \ref{fig:state-space-Linda-A}. 

A different explanation is that subjects interpreted `and' to mean `either' rather than as logical conjunction.\footnote{Although not as immediately intuitive, this interpretation of `and' is also common as in ``He invited friends and colleagues to the party.'' \cite{hertwig2008conjunction} relate this to the conjunction fallacy.}
This is captured by $(\W',t')$ with $t': \tt  \mapsto \{w'_1\}, \tt\land\ff \mapsto \{w'_1, w'_2\}$, and $\ff \mapsto \{w'_2\}$, shown in Figure \ref{fig:state-space-Linda-B}. How can an experimenter distinguish between the potentially many different explanations of subjects' behavior? Notice that $t'(\tt) \subseteq t'(\tt \land \ff)$ while $t(\tt) \centernot{\subseteq} t(\tt \land \ff)$, so that subjects with IOU $(W',t')$ will perceive that $\tt$ implies $\tt \land \ff$ while those with IOU $(W,t)$ will not. As such, these explanations are distinguished by subjects' propensity to bet on $\{\tt \land \ff \}$ versus $\{\tt \land \ff, \tt\}$.
\hfill $\square$
\end{example}

Next, in Section \ref{sec:updating}, we examine the dynamics of contingent thinking. Without the structure imposed by an objective state-space, beliefs can be updated in novel ways that find no counterpart in the fully rational model. In particular, we distinguish between two kinds of updating: (i) \emph{event-based} updating, where the \dm learns an event, that some subset of the state-space is true, and (ii) \emph{implication-based} updating, where the \dm learns an implication, that the relation `if  $\phi$ then $\psi$' is true, which may not correspond to any subset of the state-space. 

Event-based updating---reducing the state-space by ruling out potential resolutions of uncertainty---is the canonical updating of the rational model. Theorem \ref{thm:standardUpdating} characterizes event-based updating via dynamic consistency requirements on the \dm's dynamic preferences over bets. Apposite to recent experimental findings, a corollary to this result shows that event-based updating \emph{always} leads to better contingent reasoning.\footnote{E.g., \cite{esponda2014hypothetical,martinez2019failures,enke2020you,park2023complexity}.}

Although learning an event can result in learning novel implications,\footnote{For example, precipitation does not in general imply snow, but after learning that the temperature is sufficiently cold, the implication holds} in the rational model, implications are only ever learned subordinate to learning some fact/event.  When her initial contingent thinking is flawed, however, the \dm may come to `realize' novel implications that are not precipitated by the arrival of any external information. Theorem \ref{thm:qStandad} weakens dynamic consistency to captures implication-based updating, and shows that all such `realizations' correspond to specific transformations of the \dm's IOU. 

The distinction between these two forms of updating serves as a new inroad for investigating how and why subjects improve their contingent thinking. Specifically, by characterizing updating in terms of betting behavior, we provide a direct way of distinguishing between improvements in contingent thinking that result from externally resolving uncertainty and improvements that result from introspection or from reframing the decision problem. Understanding what drives improvements in reasoning could allow for designing experiments, mechanisms, and contracts that are less prone to misinterpretation and confusion.
The following example illustrates the difference between event- and implication-based updating.

\addtocounter{example}{-1}
\begin{example}[continued]
Consider the \dm whose preferences are initially represented by the IOU presented in Figure \ref{fig:state-space-Linda-A}. 
Suppose now the \dm realizes that (oh, of course!) $\tt \land \ff$ implies $\tt$ (and also $\ff$).
According to her initial understanding, there are no non-trivial implications between the primitive statements, $\tt$, $\ff$, and $\tt \land \ff$: for any distinct $\phi, \psi \in \{ \tt, \ff, \tt \land \ff\}$, the \dm would strictly prefer to bet on $\{\phi,\psi\}$ than on $\{\phi\}$.

After realizing that $\tt \land \ff$ implies $\tt$, the \dm will entertain a new preference over bets, becoming indifferent between betting on $\{\tt, \tt \land \ff\}$ and $\{\tt\}$ (and analogously for $\ff$). This new preference can be represented by an IOU with the same state-space, and where truth assignments grow monotonically (additions to the truth assignment in bold):
$$
t^\circ: 
\qquad 
\tt  \mapsto \{\w_1, \bm{w_2}\}, \quad
\tt\land\ff \mapsto \{\w_2\}, \quad
\ff \mapsto \{\bm{\w_2}, \w_3\}.
$$
This updated IOU corresponds to the standard (i.e., objective) interpretation, as shown in Figure \ref{fig:state-space-Linda-Rat}. Despite `learning' new implications, no states have been eliminated; this updating procedure does not arise as a consequence of learning an event.
\hfill $\square$
\end{example}

Finally, in Section \ref{sec:SIDEU}, we demonstrate the connection between flawed contingent reasoning and violations of state-wise dominance. We consider more complex \emph{acts}, choice objects that yield many distinct outcomes. To describe an act to the \dm, the modeler must dictate, using linguistic statements, how the outcome of the act depends on the resolution of uncertainty. 
When the modeler and \dm entertain differing interpretations of statements, the \dm may not be able to uniquely translate this description into her subjective state-space. Specifically, there may be states where the \dm cannot determine which of the possible outcomes will be realized, yielding multiple potential translations of the act as functions over her state-space.

To account for this linguistic ambiguity, we introduce a decision-making criterion, \emph{Sparse Interpretation-Dependent Expected Utility} (SIDEU). A SIDEU \dm acts as cautiously as possible: she prefers $f$ to $g$ only when every translation of $f$ yields a higher expected payoff than every translation of $g$.
Reminiscent of \emph{obvious dominance} or \emph{twofold multiprior preferences}, SIDEU makes explicit the intuition that errors in contingent reasoning give rise to imprecise valuations of acts.\footnote{See for example \cite{li2017obviously,zhang2017partition,echenique2022twofold}.} 
As illustrated by the following example, the property of state-wise dominance (but not obvious dominance) can be lost in translation.

\addtocounter{example}{-1}
\begin{example}[continued]
Suppose that the \dm is presented with the additional statement:
\begin{itemize}[leftmargin=2cm, labelsep=.2em, itemsep=-.7ex]
\small
\item[$\neg \tt =$]``Linda is not a bank teller.''
\end{itemize}
Assume the \dm's understanding is given by $(W,t)$ from Figure \ref{fig:state-space-Linda-A}, with the addition that $\neg \tt$ is understood to be the negation of $\tt$: $t(\neg \tt) = W - t(\tt) = \{\w_2,w_3,w_4\}$.

\begin{figure}[]
  \centering
  \begin{subfigure}[t]{0.45\textwidth}{}
    \centering
    \begin{tikzpicture}[
                every node/.style   ={font=\footnotesize},
        gridlines/.style    ={gray,dotted,thin},
        x=1.3cm,
        y=0.3cm
    ]
      \def\n{4}
      \def\m{4}

      \foreach \x in {-1,...,\numexpr\n-1}{
          \draw[gridlines] (\x+1,-1) -- (\x+1,\m+.5);
      }
      \foreach \y in {-1,...,\numexpr\m-1}{
          \draw[gridlines] (0,\y+1) -- (\n+.3,\y+1);
      } 

      \draw[ultra thick,lam2] (0,3) -- (2,3);
      \draw[ultra thick,lam2] (2,1) -- (4,1);

      \draw[ultra thick,lam1] (0,4) -- (2,4);
      \draw[ultra thick,lam1] (2,2) -- (4,2);

      \draw[very thick] (0,-1.25) -- (0,\m+0.5);
      \draw[very thick] (-0.25,0) -- (\n+0.25,0);

      \foreach \i in {0,...,\numexpr\n-1}{
           \node[above=4pt] at (\i+0.5,-2.1) {$w_{\number\numexpr\i+1}$};
      }

      \foreach \i in {1,...,4}{
           \node[above=4pt] at (-.2,\i-1.2) {\scriptsize $\number\numexpr\i$};
      }

    \tikzset{fixedlabel/.style={below, align=center}}

  \node[fixedlabel] at (0+0.5,-1) {$\ \ \ \tt$ \phantom{$\land$}};
  \node[fixedlabel] at (1+0.5,-1) {$\tt,\ff, \tt {\land} \ff$};
  \node[fixedlabel] at (2+0.5,-1) {$\ \ \ \ff, \neg \tt$\phantom{$\land$}};
  \node[fixedlabel] at (3+0.5,-1) {$\ \neg\tt$\phantom{$\land$}};

    \end{tikzpicture}
    \caption{$\actF$ and $\actG$ according to modeler}
    \label{fig:actA}
  \end{subfigure}
  \hfill
  \begin{subfigure}[t]{0.45\textwidth}{}
    \centering
    \begin{tikzpicture}[
                every node/.style   ={font=\footnotesize},
        gridlines/.style    ={gray,dotted,thin},
        x=1.3cm,
        y=0.3cm
    ]
      \def\n{4}
      \def\m{4}

      \foreach \x in {-1,...,\numexpr\n-1}{
          \draw[gridlines] (\x+1,-1) -- (\x+1,\m+.5);
      }
      \foreach \y in {-1,...,\numexpr\m-1}{
          \draw[gridlines] (0,\y+1) -- (\n+.3,\y+1);
      }


      \draw[ultra thick,lam2] (0,3) -- (1,3);
      \draw[ultra thick,lam2] (2,1) -- (4,1);

        \fill[pattern={Lines[angle=45,distance={1.2pt},line width={0.6pt}]}, pattern color=lam2!20, draw=none] (1,1) rectangle ++(1,2);

      \draw[ultra thick,lam2] (1,3) -- (2,3);
      \draw[ultra thick,lam2] (1,1) -- (2,1);

      \draw[ultra thick,lam1] (0,4) -- (1,4);
      \draw[ultra thick,lam1] (1,2) -- (4,2);

      \draw[very thick] (0,-1.25) -- (0,\m+0.5);
      \draw[very thick] (-0.25,0) -- (\n+0.25,0);

      \foreach \i in {0,...,\numexpr\n-1}{
           \node[above=4pt] at (\i+0.5,-2.1) {$w_{\number\numexpr\i+1}$};
      }

      \foreach \i in {1,...,4}{
           \node[above=4pt] at (-.2,\i-1.2) {\scriptsize $\number\numexpr\i$};
      }

    \tikzset{fixedlabel/.style={below, align=center}}

  \node[fixedlabel] at (0+0.5,-1) {$\ \ \ \tt$ \phantom{$\land$}};
  \node[fixedlabel] at (1+0.5,-1) {$\tt {\land} \ff, \neg \tt$};
  \node[fixedlabel] at (2+0.5,-1) {$\ \ \ \ff, \neg \tt$\phantom{$\land$}};
  \node[fixedlabel] at (3+0.5,-1) {$\ \neg\tt$\phantom{$\land$}};

    \end{tikzpicture}
    \caption{$\actF$ and $\actG$ according to \dm}
    \label{fig:actB}
  \end{subfigure}

  \caption{Acts $\actF$ and $\actG$. For the \dm, the act $\actG$ is ambiguous, taking multiple values in $\w_2$. The statements listed beneath each state are those modeled as true in that state.}
  \label{fig:acts-pair}
\end{figure}

A \emph{syntactic act} is a description of an action, a collection of bets that specify different payoffs. Consider two acts: act $\actF$ pays 4 if $\tt$ is true and pays 2 if $\neg \tt$ is true; act $\actG$ pays 3 if either $\tt$ or $\tt \land \ff$ is true and pays 1 if $\neg \tt$. That is:
$$
\actF = \Big\{ 
\begin{tabular}{r l}
$\tt $  & \hspace{-2ex} $\mapsto 4$ \\[-2ex]
$\neg \tt$  & \hspace{-2ex} $\mapsto 2$
\end{tabular}
\qquad\qquad
\actG = \Big\{ 
\begin{tabular}{rl}
$\{\tt,\tt\land\ff\} $  & \hspace{-2ex} $\mapsto 3$ \\[-2ex]
$\neg \tt$  & \hspace{-2ex} $\mapsto 1$
\end{tabular}
$$
From a (logically sound) modeler's perspective, embodied by the IOU $(W,t^\circ)$ in Figure \ref{fig:state-space-Linda-Rat}, $\tt \land \ff$ implies $\tt$. As such, these two acts are predicated on the same partition of the state-space, and moreover, $\actF$ state-wise dominates $\actG$, as shown in Figure \ref{fig:actA}. From the \dm's perspective, however, the act $\actG$ is ambiguous: both the statements $\tt \land \ff$ and $\neg \tt$ are true in state $\w_2$, and so, the outcome of the action is not uniquely defined. While the worst-case translation of $\actG$ into a function over $W$ (given by the lower envelope) is state-wise dominated by $\actF$, the best-case translation (given by the upper envelope) is not. So, a \dm who finds $\w_2$ sufficiently likely cannot rule out the possibility that $\actG$ yields a higher expected utility than $\actF$, justifying $\actG$'s choice. Nonetheless, if $\actF$ obviously dominated $\actG$, then no translation would justify its choice.
\hfill $\blacksquare$
\end{example}

The remainder of the paper is structured as follows. Section \ref{sec:prelim} introduces the domain, and contains our analysis of subjective implication: characterization, representation, and identification results. Section \ref{sec:updating} contains the analysis of updating and Section \ref{sec:SIDEU} the application to violations of dominance. 
A literature review can be found in Section \ref{sec:lit}.
 The \hyperref[sec:proofs]{Appendix} contains all proofs and auxiliary results.

\section{Identifying Subjective Implication} 
\label{sec:prelim}

\def\cont{contingency\xspace}
\def\conts{contingencies\xspace}

We begin with a language $\L$, which is an arbitrary set of payoff-relevant \textit{statements}. We think of these statements as real linguistic ones, expressed in human language, for example ``It is raining'' or ``The S\&P 500 went up today.'' For now, we impose no restrictions on $\L$, but will later (in Section \ref{sec:structure}) consider additional structural assumptions on the relation between statements.
Let $\K$ collect the set of all nonempty subsets of $\L$, which we will refer to as \textit{\conts}. We use lowercase $\phi,\psi,...$ for generic elements of $\L$, and uppercase $\Phi,\Psi,...$ for those of $\K$.

To understand how a \dm interprets uncertainty, the modeler must observe her choices regarding syntactic objects, actions that relate payoffs to the statements in $\L$. We take as the primitive the \dm's preferences over \emph{bets}, syntactic acts which pay a fixed payoff contingent on a set of winning conditions.\footnote{Later, in Section \ref{sec:SIDEU}, we investigate how a \dm's interpretation of uncertainty affects her decisions over more complex choice objects which have possibly many distinct outcomes.} Formally, a (syntactic) bet is a pair, $x_\Phi = (x,\Phi)$, where $x \in \R_+$ is a payoff and $\Phi \in \K$ is a \cont.
This way, $x_\Phi$ represents a bet that pays $x \geq 0$, denominated in utils, should any of the statements $\phi \in \Phi$ turn out to be true, and pays 0 otherwise.
The observable of the model is a preference relation $\s$ on the set of bets.%

\subsection{Subjective Implication} 
\label{sec:si}

\def\p{\piv}
\def\q{\textsc{q}}

When observing a \dm's choices in a decision problem, a modeler will want to make inferences about the \dm's understanding of specific contingencies and her ability to reason about the relation between them. We here posit a behavioral definition of implication that directly regards the primitive observable data. As such, this definition is universally applicable to any larger model of decision-making and is agnostic to any further assumptions about the \dm's behavior.

To gather intuition for this central concept, consider a \dm who is indifferent between betting on $\phi$ alone or on betting on both $\phi$ and $\psi$: $x_{\{\psi\}} \sim x_{\{\phi, \psi\}}$.
The \dm's indifference---her belief that the additional criterion for receiving $x$ contingent on $\phi$ has no added benefit---exposes her understanding that whenever $\phi$ is true, $\psi$ is also true. In other words, that $\phi$ implies $\psi$. Notice that if there were some other context (i.e., winning conditions) in which $\phi$ did add value beyond $\psi$, then this would falsify the \dm's belief that $\phi$ implies $\psi$, and so, our general definition of implication requires that this relation holds also in the presence of any other winning conditions $\Gamma$.

\begin{definition}[Subjective Implication]
\label{def:si}
For two \conts, $\Phi, \Psi \in \K$, say that a \dm with preference $\s$ perceives that \df{$\Phi$ implies $\Psi$}, written $\Phi \imp \Psi$,  if 
$$x_{\Psi \cup \Gamma} \sim x_{\Phi \cup \Psi  \cup \Gamma}$$
for all $x \in \R_+$ and 
$\Gamma \in \K$.
\end{definition}


  Given sufficient structure, 
  a bet on the contingency $\Phi =  \{\phi_1, \ldots, \phi_n\}$ will be seen as equivalent to a bet on the single statement $\phi$ = ``$\phi_1  \lor  \ldots \lor  \phi_n$'' (i.e., the statement ``$\phi_1  \textbf{\texttt{ or }} \ldots \textbf{\texttt{ or }} \phi_n$''). Thus, our explicit decision to model implication over contingencies, rather than only over specific statements, is because we do not want to universally impose structural demands on the language
  or on the \dm correctly associating a \textit{disjunction} to the set of winning conditions it represents. (In Section \ref{sec:structure} we characterize this additional requirement.)

Two important properties that arise from the implication operator are subjective \emph{nullness } and \emph{disjointness}:

\begin{definition}[Nullness and Disjointness]
\label{def:NullDis}
For two \conts, $\Phi, \Psi \in \K$ we say that the \dm perceives that:
\begin{itemize}[itemsep=-.5ex]
   \item $\Phi \in \K$ is \df{null} if $\Phi \imp \Gamma$ for all $\Gamma \in \K$;
   \item $\Phi$ and $\Psi$ are \df{disjoint}, written $\Phi \bot \Psi$, if $\Gamma \imp \Phi$ and $\Gamma \imp \Psi$ implies that $\Gamma$ is null.
 \end{itemize}
\end{definition}

\emph{Null} \conts are those that are perceived to trivialize the interpretation, in that if they were true, everything else would follow. In classical models of logical inference, $\Phi$ is null if and only if it is a contradiction, that is, if it is necessarily false. Accordingly, two statements are perceived to be \emph{disjoint} if there are no non-null \conts that jointly imply them; in other words, $\Phi$ and $\Psi$ are perceived as disjoint if they can only simultaneously hold in the trivial state-of-affairs where everything is true. 

\begin{example}
\label{ex:2}
There are four statements $\L = \{\aa,\bb,\cc,\dd\}$. The DM's preference is such that there are four $\s$-indifference-classes
\begin{align*}
A &= \big\{ \{\aa\}, \{\aa, \bb\}, \{\aa, \cc\}, \{\aa, \dd\}, \{\bb,\cc\}, \{\aa, \bb, \cc\},  \\[-1.5ex]
&\qquad\qquad  \{\aa, \bb, \dd\},  \{\aa, \cc, \dd\},  \{\aa, \bb, \cc, \dd\},  \{\bb,\cc,\dd\} \big\} \\
B &= \big\{ \{\bb\}, \{\bb,\dd\}  \big\},\qquad  C = \big\{ \{\cc\},  \{\cc,\dd\}\big\} \qquad D= \big\{ \{\dd\}  \big\}
\end{align*}
such that $A \succ B \succ D$ and $A \succ C \succ D$ (and $B$ and $C$ are incomparable). Using the definition of subjective implication, it is easy to see that these are also $\imp$-equivalence classes and that $D \imp A$, $D \imp B$, $D \imp C$, $B \imp A$, and $C \imp A$, as well as all reflexive relations (i.e., the implication holds for any members of the specified equivalence classes; e.g., $\{\cc,\dd\} \imp \{\bb,\cc\}$ since $\{\bb,\cc\} \sim \{\bb,\cc\} \cup \{\cc,\dd\}$, etc.). 

From this we can see that elements of $D$ are null, since they imply all other statements. Likewise, elements of $B$ and $C$ are disjoint, since the only elements that imply them both are in $D$, and so null. On the other hand, elements of $B$ and $A$ are not disjoint, since $\bb$ implies them both and is not null (since $\bb$ does not imply $\dd$).
\hfill $\square$
\end{example}

\subsection{Representation}
\label{sec:iou}

It turns out that under exceedingly weak conditions on $\s$, the \dm's purely subjective notion of implication can be viewed as arising from a state-space interpretation of uncertainty. That is, there exist a state-space $\W$ and an assignment $t: \K \to 2^\W$ such that the \dm's exposed understanding of implications is represented by the state-space. Towards this: 

\begin{definition}[Interpretation of Uncertainty]
An \df{interpretation of uncertainty} (IOU) is a pair $(\W,t)$ where:
$\W$ is a set of \df{states} and $t:\K \rightarrow 2^{\W}$ is a \df{truth assignment}.
\end{definition}

An IOU, $(\W,t)$,  is a particular way of interpreting \conts and how they relate to one another. Each state $\w \in \W$ represents a possible way that uncertainty can resolve, where exactly the contingencies  $\{ \Phi \in \K \mid \w \in t(\Phi)\}$ are true. 
Thus, identifying how the \dm  understands the relation between statements is tantamount to identifying an IOU that faithfully captures her preferences.

\begin{definition}[Faithfulness]
Say that an IOU \df{is faithful to} $\s$ if 
\begin{enumerate}[label=\textup{(F\arabic*)},leftmargin=0.5in,itemsep=-.5ex]
\item\label{faithful:subset}$t(\Phi) \subseteq t(\Psi)$ if and only if $\Phi \imp \Psi$,
\item\label{faithful:null}$t(\Phi) = \emptyset$ if and only if $\Phi$ is null, and
\item\label{faithful:disjoint} $\w \in t(\Phi) \cap t(\Psi)$ implies $\w \in t(\Gamma)$ for $\Gamma \in \K$ such that  $\Gamma \imp \Phi$ and $\Gamma \imp \Psi$.
\end{enumerate}
\end{definition}

A faithful interpretation is one that accurately reflects the \dm's revealed subjective understanding of implications. The first dictate of faithfulness requires that the \dm believes that $\Phi$ implies $\Psi$ exactly when there is no state where $\Phi$ holds and $\Psi$ does not, reflecting our colloquial understanding of implications. The other two dictates require that properties of nullness and disjointness, as defined through the primitive, are also characterized by their set-theoretic analogs.\footnote{Note that \ref{faithful:disjoint} implies $t(\Phi) \cap t(\Psi) = \emptyset$ if and only if $\Phi\bot\Psi$. It is in fact slightly stronger than this, also requiring that \emph{every} state where both $\Phi$ and $\Psi$ hold witnesses their non-disjointness.}

Given the complete lack of structure on $\L$, finding a faithful interpretation seems an onerous request. At first glance, there is little reason to believe that a \dm's abstract understanding of statements, as revealed by her betting behavior, can be so succinctly embodied via a state-space model. Contrary to our intuition, however, it turns out that the requirements on $\s$ to ensure a faithful interpretation are extremely weak.\footnote{As noticed by a referee, the necessary conditions are even weaker: as the definition of subjective implication speaks only of indifference relations, \emph{strict} transitivity is not necessary.}

 \begin{theorem}
 \label{thm:iounoprob}
 If $\s$ is reflexive and transitive, then there exists an IOU faithful to $\s$.
 \end{theorem}

The state-space is formed by collecting suitably structured filters of the subjective implication relation, and as such, the proof of Theorem \ref{thm:iounoprob} follows the strategy of many classic results in algebraic logic \`a la Stone's duality. 
Each state represents a 
resolution of uncertainty according to the \dm: it is identified with a set of mutually consistent statements (i.e., those considered true at the state).
The departure from classical results is that `consistent' is w.r.t.~the subjective implication, rather than an external logic. This construction is explored in the following example.

\begin{figure}[]
  \centering
  \begin{subfigure}[t]{0.3\textwidth}{}
    \centering
    \begin{tikzpicture}[
  every node/.style={draw, circle, minimum size=4.3mm, inner sep=0pt, font=\scriptsize},
  >=Stealth,
  line width=1pt,    
  shorten >=1pt, shorten <=1pt,
  scale=.7
]

 \definecolor{tt}{HTML}{c45b5b}
  \definecolor{ff}{HTML}{5186b5}
  \definecolor{tf}{HTML}{d4ad00}

  \node[fill=tf!50, ] (A) at (0,  1) {$A$};
  \node[fill=tt!50, ] (B) at (-1, 0) {$B$};
  \node[fill=ff!50, ] (C) at ( 1, 0) {$C$};
  \node (D) at (0, -1) {$D$};

  \draw[->] (D) -- (B);
  \draw[->] (D) -- (C);
  \draw[->] (B) -- (A);
  \draw[->] (C) -- (A);
\end{tikzpicture}

    \caption{Hasse diagram for $\imp$.}
    \label{fig:imp_graph}
  \end{subfigure}
  \hfill
  \begin{subfigure}[t]{0.3\textwidth}
    \centering
    \begin{tikzpicture}[
        every node/.style   ={font=\footnotesize},
        gridlines/.style    ={gray,dotted,thin},
        x=.9cm,
        y=0.4cm
    ]
      \def\n{3}
      \def\m{3}

      \definecolor{tt}{HTML}{c45b5b}
      \markcells{2/1}{tt!50}{$B$}
      \markcells[pattern={Lines[angle=45,distance={1.2pt},line width={0.6pt}]}, pattern color=tt!30]{1/1, 3/1}{}{}

      \definecolor{ff}{HTML}{5186b5}
      \markcells{3/2}{ff!50}{$C$}
      \markcells[pattern={Lines[angle=45,distance={1.2pt},line width={0.6pt}]}, pattern color=ff!30]{1/2, 2/2}{}{}

      \definecolor{tf}{HTML}{d4ad00}
      \markcells{1/3,2/3,3/3}{tf!50}{$A$}

      \foreach \x in {-1,...,\numexpr\n-1}{
          \draw[gridlines] (\x+1,-1) -- (\x+1,\m+.5);
      }
      \foreach \y in {-1,...,\numexpr\m-1}{
          \draw[gridlines] (-.3,\y+1) -- (\n+.3,\y+1);
      }

      \draw[thick] (0,0) rectangle (\n,\m);
      \foreach \i in {1,...,\numexpr\n-1}{
          \draw (\i,0) -- (\i,\m);
      }

      \draw[very thick] (-0.25,0) -- (\n+0.25,0);

          \node[above=4pt] at (0+0.5,-1.7) {$w_{A}$};
          \node[above=4pt] at (1+0.5,-1.7) {$w_{B}$};
          \node[above=4pt] at (2+0.5,-1.7) {$w_{C}$};
    \end{tikzpicture}
    \caption{IOU $(\W,t)$}
    \label{fig:ex-A}
  \end{subfigure}
  \hfill
  \begin{subfigure}[t]{0.3\textwidth}{}
    \centering
    \begin{tikzpicture}[
                every node/.style   ={font=\footnotesize},
        gridlines/.style    ={gray,dotted,thin},
        x=.9cm,
        y=0.4cm
    ]
      \def\n{2}
      \def\m{3}

    \definecolor{tt}{HTML}{c45b5b}
      \markcells{1/1}{tt!50}{$B$}
      \markcells[pattern={Lines[angle=45,distance={1.2pt},line width={0.6pt}]}, pattern color=tt!30]{2/1}{}{}

      \definecolor{ff}{HTML}{5186b5}
      \markcells{2/2}{ff!50}{$C$}
      \markcells[pattern={Lines[angle=45,distance={1.2pt},line width={0.6pt}]}, pattern color=ff!30]{1/2}{}{}

      \definecolor{tf}{HTML}{d4ad00}
      \markcells{1/3,2/3}{tf!50}{$A$}

      \foreach \x in {-1,...,\numexpr\n-1}{
          \draw[gridlines] (\x+1,-1) -- (\x+1,\m+.5);
      }
      \foreach \y in {-1,...,\numexpr\m-1}{
          \draw[gridlines] (-.3,\y+1) -- (\n+.3,\y+1);
      }

      \draw[thick] (0,0) rectangle (\n,\m);
      \foreach \i in {1,...,\numexpr\n-1}{
          \draw (\i,0) -- (\i,\m);
      }

      \draw[very thick] (-0.25,0) -- (\n+0.25,0);

      \foreach \i in {0,...,\numexpr\n-1}{
           \node[above=4pt] at (\i+0.5,-1.7) {$w_{\number\numexpr\i+1}$};
      }
    \end{tikzpicture}
    \caption{IOU $(\W',t')$}
    \label{fig:ex-B}
  \end{subfigure}
  \caption{The implication relation from Example \ref{ex:2} and two IOUs faithful to it.}
  \label{fig:2}
  \end{figure}

\addtocounter{example}{-1}
\begin{example}[continued]
\label{ex:construct}
Recall the implication operator $\imp$ from earlier in the example and shown in Figure \ref{fig:imp_graph}. If $(W,t)$ is faithful to $\s$, then for any two contingencies $\Phi$ and $\Psi$,  $t(\Psi) \subseteq t(\Phi)$ exactly when there is a path from (the equivalence class containing) $\Phi$ to (the equivalence class containing) $\Psi$ in the diagram in Figure \ref{fig:imp_graph}.

When $\L$ is finite, finding a faithful IOU is straightforward. For each equivalence class of non-null contingencies we create a state, and in that state, a contingency is true iff it is implied by the contingency that defines the state. So, in our example, we have three states, indexed by $A$, $B$, and $C$. In $w_A$, a contingency is true iff it is implied by the elements of $A$ (this is the elements of $A$ themselves); in $w_B$ a contingency is true iff it is implied by the elements of $B$ (this is $A \cup B$); in $w_C$ a contingency is true iff it is implied by the elements of $C$ (this is $A \cup C$). This is the IOU in Figure \ref{fig:ex-B}.
\hfill $\blacksquare$
\end{example}

\subsection{Uniqueness}
\label{sec:ident}
\newcommand{\exxt}[1]{\langle#1\rangle}


A faithful interpretation is an understanding of the world that generates the \dm's behavior; each state captures a possible resolution of uncertainty (i.e., determines exactly which \conts are true). As such, identifying a \dm's interpretation is tantamount to identifying exactly how she believes uncertainty might resolve. In this section, we will show that under weak conditions there exists a maximal and a minimal interpretation; the maximal state-space captures those resolutions of uncertainty which are \emph{consistent with} the \dm's preferences, and the minimal state-space those \emph{necessary to explain} her preferences. We further show that any intermediate state-space is also a faithful interpretation.

\begin{definition}
Let $(\W,t)$ and $(\W',t')$ be two IOUs.
Say that $(\W',t')$ \df{extends} $(\W,t)$ if for all $\w \in \W$ there is some $\w' \in \W'$ such that 
$$\{\Phi \in \K \mid \w \in t(\Phi) \} = \{\Phi \in \K \mid \w' \in t'(\Phi) \}.$$
\end{definition}


One interpretation extends another if every state in the latter can be thought of as some state in the former. That is, up to relabeling, the extension simply adds additional states. The IOUs that are faithful to some $\s$ are highly structured, allowing a modeler to meaningfully identify the uncertainty perceived by the \dm. 

\begin{theorem}
 \label{thm:maxstatespace}
Let $\s$ be reflexive and transitive. Then:
\begin{enumerate}
  \item There exists a maximal faithful IOU of $\s$, i.e., that extends any other faithful IOU. 
  \item If the set of $\imp$-equivalence classes is finite, then there also exists a
   minimal faithful IOU of $\s$, i.e., that is extended by any other faithful IOU. 
\item If $(\W,t)$ both extends and is extended by faithful IOUs, it is itself a faithful IOU.
\end{enumerate}
Also, the maximal and minimal IOUs are unique up to duplicating or relabeling states.
 \end{theorem}

Each state in a faithful IOU determines a potential resolution of uncertainty, as it specifies a collection of contingencies that could be simultaneously true. As such, duplicating a state (or renaming it) does not change \emph{which} resolutions of uncertainty are considered possible. Maximal and minimal IOUs, therefore, identify the set of possible resolutions of uncertainty that might be considered by the \dm: A maximal IOU contains all resolutions of uncertainty that are consistent with the \dm's betting behavior (i.e., if $(W,t)$ contains a state $\w$ such that the collection of contingencies true at $\w$ does not coincide with those true at some state in the maximal IOU, then $(\W,t)$ cannot be faithful to $\s$). Conversely, a minimal IOU captures those resolutions that \emph{must} be present to explain the \dm's behavior.

The final claim also shows that these bounds on the set of faithful interpretations are the best that we can hope for;  if we found two IOUs, one extending the other, then \emph{any} set of states contained in the larger, but not the smaller, could be removed without behavioral consequence.

The faithful IOU in Figure \ref{fig:ex-A} is not minimal. Indeed, notice that the IOU in Figure \ref{fig:ex-B} is extended by it (via the map $w_1 \mapsto w_B$ and $w_2 \mapsto w_C$); this latter IOU turns out to be minimal (and the former is maximal, up to an `empty-state' where no statements are true). The construction provided earlier, in Example \ref{ex:construct} will always pick out the maximal faithful interpretation. In the proof of Theorem \ref{thm:iounoprob}, we provide a more technical construction of \emph{canonical} state-spaces, that turn out to be minimal for finite languages.

\subsection{Toolbox for Identification}
\label{sec:structure}
The model so far does not specify any relationship between contingencies, and therefore cannot determine when contingent thinking succeeds or fails. To account for empirical failures of logical thinking, then, requires two ingredients: first, delineating what rational behavior \emph{would} entail, to understand what it means to deviate from it, and second, adding structure to the language, to be able to classify behavior along structural lines. To specify rational behavior, we take as given the \emph{modeler's implication} operator; we write $\phi \implies \psi$ to mean that, under the modeler's interpretation of statements, $\phi$ implies $\psi$. Take as shorthand $\phi \iff \psi$ to mean that $\phi \implies \psi$ and $\psi \implies \phi$, i.e., that $\psi$ and $\phi$ are logically equivalent. 

To the second point, we now assume the language is closed under several connectives. In particular, if $\phi,\psi$ are in $\L$ then we assume also the existence of the statements: $\neg \phi$, $(\phi \land \psi)$, and $(\phi \lor \psi)$. The interpretation is the usual one: $\neg \phi$, the \emph{negation} of $\phi$, is interpreted as the statement that $\phi$ is not true, $\phi\land\psi$, the \emph{conjunction} of $\phi$ and $\psi$, is interpreted as the statement that both $\phi$ and $\psi$ are true, and $\phi\lor\psi$, the \emph{disjunction} of $\phi$ and $\psi$, is interpreted as the statement that at least one of $\phi$ and $\psi$ is true. An IOU $(W,t)$ is \emph{sound} if it understands these logical connectives in accordance with their interpretations in classical logic, so that $t$ maps conjunctions to intersections, disjunctions to unions, and negation to complements (i.e., $t$ is a Boolean homomorphism); see for example, Section 5 of \cite{sikorski1969boolean}.\footnote{While we do not require it, it is often convenient to think of $\implies$ as resulting from some traditional (classical) logical deduction and therefore results also in a sound interpretation.} 

Many empirical failures of contingent thinking are the direct result of the \dm systematically misinterpreting these logical connectives. For example, \cite{tversky1983extensional} finds subjects misinterpreting conjunction (see Example \ref{ex:kt}), \cite{tversky1994support}, disjunction, and \cite{tversky1981framing}, negation. Note, not all failures arise this way, as even a sound \dm can still interpret contingencies in contradiction to the modeler's view of implication. Example \ref{ex:ev}, below, considers subjects who fail to understand the decision environment but who are \emph{internally} consistent. 

In what follows, we provide the conditions on the \dm' preference that characterize the choice patterns predicted for different benchmarks of contingent reasoning, as captured by each of the properties of $t$. We first focus on three:

\begin{axiom}{R}{E}{Equivalence} 
If $\psi \iff \phi$ then $\psi \imp \phi$.
\end{axiom}

\vspace{-1ex}

\begin{axiom}{R}{I}{implication} 
If $\psi \implies \phi$ then $\psi \imp \phi$.
\end{axiom}

\vspace{-1ex}

\begin{axiom}{R}{C}{Conjunction}
$\Gamma \imp \phi$ and $\Gamma \imp \psi$ if and only if $\Gamma \imp \phi \land \psi$
\end{axiom}

In words, property \refax{E} characterizes a \dm that perceives objective logical equivalences.
Property \refax{I} strengthens \refax{E}, characterizing a \dm who recognizes not only equivalences but also single directional implications. Property \refax{C} captures the behavior of a \dm who understands conjunctions: if she understands that something implies both $\phi$ and $\psi$, she can conclude that they must jointly hold. Each of these properties is reflected by an analogous restriction on the set of faithful interpretations:

\begin{theorem}
\label{thm:rep-all}
Let $\s$ be reflexive and transitive. Then $\s$ satisfies 
\begin{enumerate}[label=\textup{(\arabic*)}, itemsep=-1ex]
\item \label{thm:rep-e} \refax{E} iff for all faithful $(\W, t)$ and $\phi,\psi \in \L$, $\phi \iff \psi$ implies $t(\phi) = t(\psi)$,
\item \label{thm:rep-i}  \refax{I}  iff for all faithful $(\W, t)$ and $\phi,\psi \in \L$, $\phi \implies \psi$ implies $t(\phi) \subseteq t(\psi)$,
\item \label{thm:rep-c} \refax{C} iff for all faithful $(\W, t)$ and $\phi,\psi \in \L$, $t(\phi \land \psi) = t(\phi) \cap t(\psi)$.
\end{enumerate}
\end{theorem}


While Theorem \ref{thm:rep-all} makes inroads in characterizing the sundered components of rationality, it does not yield a full characterization.
To advance further, we require the existence of an interpretation of $\s$ that treats contingencies, i.e.,  \emph{sets} of statements, as the union of their constituent parts:

\begin{definition}[Set-Consistency]
Call an interpretation $(W,t)$ \df{set-consistent} if for all contingencies $\Phi,\Psi \in \K$, we have $t(\Phi \cup \Psi) = t(\Phi )\cup t(\Psi)$.
\end{definition}

In Example \ref{ex:2}, the IOU in Figure \ref{fig:ex-B} is set consistent whereas the IOU in Figure \ref{fig:ex-A} is not  since $t(\{\bb,\cc\}) = W \neq \{w_B,w_C\} = t(\bb) \cup t(\cc)$. The existence of a set-consistent IOU is equivalent to the following richness condition on the set of statements:

\begin{definition}[Articulation]
We call $\s$ \df{articulate} if, for all $\Phi,\Psi\in\mathcal{K}$,
\begin{enumerate}[label=\textup{(A\arabic*)},leftmargin=0.75in,itemsep=-1ex]
\item \label{art1} If $\Phi\centernot{\imp}\Psi$ then there exists a nonnull $\Gamma \in\mathcal{K}$ such that $\Gamma \imp\Phi$ and $\Gamma\bot\Psi$.

\item \label{art2} For any collection $\{\Psi_i\}_{i \in \mathcal{I}} \subseteq \K$, if  $\Phi\bot \Psi_i$ for all $i \in \mathcal{I}$ then, $\Phi\bot \bigcup_{i \in \mathcal{I}} \Psi_i$.
\end{enumerate}
\end{definition}

The first point states that when the \dm believes that $\Phi$ does \emph{not} imply $\Psi$, she can `verbally' articulate why: she can point out a specific contingency $\Gamma$ that would imply $\Phi$ but is incompatible with $\Psi$. Similarly, the second point states that if the \dm believes that $\Phi$ is compatible with some contingency $\Psi$, then the \dm can articulate a specific state of affairs where both are true: whenever $\Psi$ is broken down into subsets $\{\Psi_i\}_{i \in \mathcal{I}}$, the \dm considers some subset compatible with $\Phi$. 

In addition to providing a set-consistent interpretation, articulation allows us to capture the other structural properties of $t$ in a straightforward manner. Namely:

\begin{axiom}{R}{D}{Disjunction}
$\phi \imp \Gamma$ and $\psi \imp \Gamma$ if and only if $\phi \lor \psi \imp \Gamma$
\end{axiom}

\vspace{-1.5ex}

\begin{axiom}{R}{N}{negation} 
$\phi \bot \neg\phi$ and $\Gamma \imp \{\phi, \neg\phi\}$
\end{axiom}

In a dual axiom to \refax{C}, the former captures an understanding of disjunctions. The latter requires the \dm to understand negations---that in every possible resolution of uncertainty, either a statement or its negation (but not both) is true. We then have:

\begin{theorem}
\label{thm:rep-art}
Let $\s$ be reflexive, transitive, and articulate. Then there exists a faithful and set-consistent interpretation of $\s$. Moreover, $\s$ satisfies
\begin{enumerate}[label=\textup{(\arabic*)},itemsep=-1ex]
\addtocounter{enumi}{3}
\item \label{thm:rep-d} \refax{D} iff for all faithful, set-consistent $(\W, t)$ and $\phi,\psi \in \L$, $t(\phi \lor \psi) = t(\phi) \cup t(\psi)$,
\item \label{thm:rep-n}  \refax{N}  iff for all faithful, set-consistent $(\W, t)$ and $\phi \in \L$, $t(\neg \phi) = \bigcup_{\Phi \in \K}t(\Phi) - t(\phi)$.
\end{enumerate}
\end{theorem}

Where our general model provides a universal theory of `failures of contingent thinking,' the axioms in this section serve as further, falsifiable specifications of it. Theorems \ref{thm:rep-all} and \ref{thm:rep-art} serve the modeler with a diagnostic tool for uncovering the particular driving force behind observed deviations from rationality, allowing for the identification of \textit{types} of \dms according to the specific logical inferences they fail to internalize.

\section{Dynamic Preferences and Information Acquisition}
\label{sec:updating}

We now assume that the modeler observes a pair of (reflexive and transitive) preference relations $(\s, \ss)$ where $\s$ denotes the DM's initial preferences and $\ss$  those observed at a later time. We provide conditions so that the change in preferences from $\s$ to $\ss$ can be understood as arising from the \dm updating her interpretation of uncertainty.

The standard notion of updating in the rational model is captured by the elimination of states that are incompatible with what has been learned. We call this \emph{event-based} updating, since the \dm learns an event in the state-space. We will show that without logical rationality, there is another form of updating, which we call \emph{implication-based} updating, that is not event-based, but rather hinges on realizing novel relations between contingencies. 

\subsection{Event-Based Updating}
Event-based updating can be represented via:

\begin{definition}[Event-Based Updating]
Call $(\s, \ss)$ \df{event-based} if there exists some $(\W,t)$ and $\W^{\circ}\subseteq \W$ such that

\begin{itemize}
  \item $(W,t)$ is an interpretation of $\s$, and
  \item $(\W^{\circ}, \Psi \mapsto t(\Psi)\cap \W^{\circ})$ is a faithful interpretation of $\ss$.
\end{itemize}
\end{definition}

When $(\s, \ss)$ is event-based, it is as if the \dm's initial understanding is embodied by $(W,t)$ and she updates her interpretation by learning the event $\W^{\circ}$, and so discarding any state contained in its complement; her new truth assignment is simply the restriction of her old truth assignment to this smaller state-space.

Within our framework, event-based updating is captured by two dynamic consistency axioms.\footnote{We thank an anonymous referee for suggesting this line of inquiry.} The first states that all implications are preserved:

\begin{axiom}{}{DC-A}{Preservation of Implication} 
If $\Psi \imp \Lambda$ then $\Psi \impp \Lambda$.
\end{axiom}

That is, after updating, the \dm might believe new implications hold but cannot forget implications she already believed. 
The second dynamic consistency axiom relates to the preservation of intersections. Notice, if $\Phi$ and $\Psi$ are initially considered disjoint (their truth assignments have empty intersection), then event-based updating will maintain their disjointness.  \refax{DC-B}, below, strengthens this to further ensure well-behaved non-empty intersections:

\begin{axiom}{DCB}{DC-B}{Preservation of Intersections} 
Let $\Psi \impp \Gamma$, $\Psi \impp \Gamma'$ and $\Psi \notimpp \Lambda$. Then there exists some $\Psi'$ with $\Psi' \impp \Psi$ and such that $\Psi' \imp \Gamma$, $\Psi' \imp \Gamma'$, $\Psi' \notimpp \Lambda$.
\end{axiom}

These two axioms are necessary for event-based updating, and, so long as there are only finitely many equivalence classes of statements, they prove sufficient as well.\footnote{Intermediate in the proof of Theorem \ref{thm:standardUpdating}, we provide a characterization of event-based updating that holds universally, but is behaviorally and economically opaque.}

\begin{theorem}
\label{thm:standardUpdating}
If $(\s, \ss)$ is event-based then it satisfies \refax{DC-A} and \refax{DC-B}. Moreover, if the set of $\imp$-equivalence classes is finite, then the converse holds.
\end{theorem}

An important, if immediate, consequence of event-based updating is that it will preserve all of the rationality benchmarks presented in Section \ref{sec:structure}. Formally:

\begin{corollary}
\label{cor:persistence}
If $(\s, \ss)$ is event-based, then if $\s$ satisfies any subset of $\{\refax{E},\refax{I},\refax{C},\refax{D},\refax{N}\}$, so too does $\ss$.
\end{corollary}

Corollary \ref{cor:persistence} follows immediately from the fact that all relevant properties of truth assignments are preserved under intersection with a subset of the state-space. Of course, the converse does not hold: it is possible that a \dm who does not satisfy some benchmark of rationality prior to updating but does satisfy it after. In particular, her preferences might be `locally' rational over the subset $W^{\circ}$.

\subsection{Implication-Based Updating}

When the \dm's initial interpretation of uncertainty is flawed, it seems reasonable that, upon reflection, she might become aware of some error in her understanding and subsequently correct for it. As shown in the example in Section \ref{ex:kt}, this can lead to updated beliefs and the adoption of novel implications in the absence of any event-based updating. At its most general, this updating procedure is captured by a pair $(\s, \ss)$ that satisfies \refax{DC-A} alone, so that we insist only that implication be preserved. As such, we call it implication-based updating.

That we refer to such a procedure as updating is substantiated by the fact that, given \refax{DC-A}, $(\s, \ss)$ can each be represented by the same state-space and truth assignments that are monotone transformations of one another. Specifically:

\begin{definition}[Implication-Based Updating]
\label{def:imp-based-updating}
Call $(\s, \ss)$ \df{implication-based} if there exists some $(\W,t)$ and 
$h: 2^\W \to 2^\W$ such that

\begin{itemize}
  \item for all $A, B \in 2^W$, $A \subseteq B$ implies $h(A) \subseteq h(B)$
  \item $(W,t)$ is an interpretation of $\s$, and
  \item $(\W^{\circ}, \Psi \mapsto h(t(\Psi))\cap \W^{\circ})$ is a faithful interpretation of $\ss$.
\end{itemize}
where $\W^{\circ} = W - \bigcap_{\Phi \in \K} h(t(\Phi))$.
\end{definition}

Notice that it is immediate that event-based updating is a special case of implication-based updating, as for an arbitrary $W^{\circ}$, we can set $h:A \mapsto A \cup (\W - W^{\circ})$. Of course, this can also be seen by the fact that implication-based updating corresponds exactly to the preservation of implication:

\begin{theorem}
\label{thm:qStandad}
$(\s, \ss)$ is implication-based if and only if it satisfies \refax{DC-A}.
\end{theorem}

Implication-based updating provides a more complete view to the preservation of rationality benchmarks, supplementing the more or less trivial Corollary \ref{cor:persistence}. It is essentially by definition that implication-based updating will maintain satisfaction of \refax{I} (and therefore \refax{E}); nonetheless, it is simple to construct examples where $(\s, \ss)$ is implication-based and yet $\s$, but not $\ss$, satisfies \refax{C} (or \refax{D}, \refax{N}). The preservation of these rationality benchmarks comes down to the `shape' of the distortion function $h$---it is precisely the monotonicity of $h$ that yields the preservation of \refax{I}; if we want some benchmark to be maintained, it suffices that $h$ preserves the corresponding set operation; e.g., for \refax{C}, $h$ preserves intersections, $h(A \cap B) = h(A) \cap h(B)$, etc.

\begin{example}
\label{ex:ev}
As an essential example of flawed contingent thinking, take the following experimental findings adapted from \cite{esponda2014hypothetical}.
The state of the world is either \red or \blue with equal probability. Each subject must cast a vote for either \red or \blue without observing the state but after observing a signal---with possible realizations \red or \blue, with accuracy $\frac23$. In addition, two computers observe the state and are programmed to follow the specific rule:
\begin{itemize}[itemsep=-.7ex]
  \item If the state is \red: vote \red
  \item If the state is \blue: vote \blue with probability $\frac23$ and \red with $\frac13$. 
\end{itemize}
If a simple majority votes for the correct state, the subject's payoff is \$20; otherwise, the payoff is \$0. Subjects are made fully aware of the rule used by the computers. It is a dominant strategy to vote \blue independently of the observed signal: for the subject's vote to affect the outcome, the computers must disagree, and hence the state must be \blue. Nonetheless, 80\% of subjects do not play strategically even after 40 rounds.

A parsimonious representation of this decision problem requires the following relevant statements:
$$
\state_c  = \text{\small ``The state is $c$."} \quad \obs_c  = \text{\small ``The observed signal is $c$."} \quad \vote_{c}  = \text{\small ``Both computers voted $c$."} 
$$
for $c \in \{\red, \blue\}$, as well as
$$
\piv  = \text{\small ``Subject's vote determines the outcome."} \quad \vote_{\textsc{s}}  = \text{\small ``Computers' votes are split."}
$$
and the conjunction of any subset of these statements.  

It seems likely that subjects understand the basic mechanics of the state and signal, e.g., that  $\state_{\red} \land \state_{\blue}$ is perceived as null, $\state_{\red}$ and $\state_{\blue}$ are perceived as disjoint, etc. In fact, we can assume that in an immediate sense, the subject understands the computers' rules so that we observe the following implication:
$\state_{\red} \imp \vote_{\red}$. 

The rules of the game require that $\piv \iff \vote_{\textsc{s}}$, from which $\piv \implies \state_{\blue}$ follows from the standard calculus of logical deduction. Critically, however, we assume this is \emph{not} observed; the subject's flawed reasoning does not lead her to this essential relationship between contingencies. Instead, the subjects views $\piv$ as completely independent from other contingencies. 
This is captured by the following IOU, $(\W,t)$, where
\begin{align*}
\W = \bigcup
\begin{cases}
\{\state_{\red }\} \times \{\vote_{\red}\} \times  \{\obs_{\red}, \obs_{\blue}\}  \times \{\piv, \textsc{ not } \p\} \\
\{\state_{\blue} \} \times \{\vote_{\red},\vote_{\blue}, \vote_{\textsc{s}}\} \times \{\obs_{\red}, \obs_{\blue}\}  \times \{\piv, \textsc{ not } \piv\}
\end{cases}
\end{align*}
and $t$ maps each primitive statement to the states that contain it and maps conjunctions to intersections, shown in Figure \ref{fig:state-space-EV}. For a fixed and independent probability of  $\piv$, $\gamma$, the probabilities of each state are given in the figure. This IOU reflects the subject's perception of the relation between the statements: $t(\state_{\red} \land \state_{\blue}) = \emptyset$ and $t(\vote_{\red}) \subseteq t(\state_{\red})$, but $t(\piv) \centernot\subseteq t(\state_{\blue})$. 

\begin{figure}

    \centering
    \begin{tikzpicture}[
        every node/.style   ={font=\small},
        gridlines/.style    ={gray,dotted,thin},
        x=.6cm,
        y=0.45cm
    ]
      \def\n{16}
      \def\m{4}

      \definecolor{cr}{HTML}{c45b5b}
       \definecolor{cb}{HTML}{5186b5}
       \definecolor{cp}{HTML}{d4ad00}

       \def\perm{11,7,9,15,   12,8,10,16,  5,3,1,13,    6,2,4,14,}
      \foreach [count=\p from 1] \o in \perm {
        \expandafter\xdef\csname pos@\o\endcsname{\p}
      }

      \foreach \s in {1,...,6,13,14}{
          \edef\pp{\csname pos@\s\endcsname}
          \markcells{\pp/2}{cb!30}{$\obs_{\blue}$}
      }
      \foreach \s in {7,...,12,15,16}{
          \edef\pp{\csname pos@\s\endcsname}
          \markcells{\pp/2}{cr!30}{$\obs_{\red}$}
      }

      \foreach \s in {1,...,12}{
          \edef\pp{\csname pos@\s\endcsname}
          \markcells{\pp/4}{cb!70}{$\state_{\blue}$}
      }
      \foreach \s in {13,...,16}{
          \edef\pp{\csname pos@\s\endcsname}
          \markcells{\pp/4}{cr!70}{$\state_{\red}$}
      }

      \foreach \s in {1,2,7,8}{
          \edef\pp{\csname pos@\s\endcsname}
          \markcells{\pp/3}{cb!50}{$\vote_{\blue}$}
      }

      \foreach \s in {3,4,9,10,13,14,15,16}{
          \edef\pp{\csname pos@\s\endcsname}
          \markcells{\pp/3}{cr!50}{$\vote_{\red}$}
      }

      \foreach \s in {5,6,11,12}{
            \edef\pp{\csname pos@\s\endcsname}
            \markcells{\pp/3}{cb!50}{}%
            \markcells[pattern={Lines[angle=45,distance={2pt},line width={1pt}]},  pattern color=cr!30]{\pp/3}{}{$\vote_{\textsc{s}}$}
      }

      \foreach \s in {1,3,...,15}{
          \edef\pp{\csname pos@\s\endcsname}
          \markcells{\pp/1}{cp!50}{$\piv$}
      }

      \foreach \s in {2,4,...,16}{
            \edef\pp{\csname pos@\s\endcsname}
            \markcells[pattern={Lines[angle=45,distance={2pt},line width={1pt}]},  pattern color=cp!30]{\pp/1}{}{}
      }

      \foreach \x in {-1,...,\numexpr\n-1}{
          \draw[gridlines] (\x+1,-1) -- (\x+1,\m+.5);
      }
      \foreach \y in {-1,...,\numexpr\m-1}{
          \draw[gridlines] (-.3,\y+1) -- (\n+.3,\y+1);
      }

      \draw[thick] (0,0) rectangle (\n,\m);
      \foreach \i in {1,...,\numexpr\n-1}{
          \draw (\i,0) -- (\i,\m);
      }

      \draw[very thick] (-0.25,0) -- (\n+0.25,0);

      \foreach \i in {0,...,\numexpr\n-1}{
           \node[above=4pt] at (\i+0.5,-1.4) {\scriptsize $w_{\number\numexpr\i+1}$};
      }

      \def\numbers{%
        $8\alpha$,$8\beta$,$2\alpha$,$2\beta$,$8\alpha$,$8\beta$,
        $4\alpha$,$4\beta$,$1\alpha$,$1\beta$,$4\alpha$,$4\beta$,
        $9\alpha$,$9\beta$,$18\alpha$,$18\beta$
      }

      \foreach [count=\orig from 1] \val in \numbers {
        \edef\pp{\csname pos@\orig\endcsname}
        \node[below=1pt] at ({\pp-0.5},-.6) {\scriptsize \val};
      }

    \end{tikzpicture}
    \caption{The IOU $(W,t)$ rationalizing subjects behavior; only the primitive statements are represented in each state. The numbers on the bottom row represent the probability of each state according to the experimental design outlined above and given a fixed and independent probability $\gamma$ of $\piv$, where $\alpha = \frac{\gamma}{108}$ and  $\beta = \frac{1-\gamma}{108}$.}
    \label{fig:state-space-EV}

\end{figure}

The subject prefers to vote red whenever she prefers a bet on $\state_{\red}\land \piv$ to a bet on $\state_{\blue} \land \piv$. To analyze this decision problem, we must consider the subject's conditional preferences: Upon observing the \red signal, that is upon learning $\obs_{\red}$ is true, her new state-space is $\W^{\obs_{\red}} = t(\obs_{\red}) = \{w_1, \ldots \w_8\}$ and her new interpretation $t^{\obs_{\red}}(\psi): \psi \to t(\psi) \cap t(\obs_{\red})$.

Notice that $t(\state_{\blue} \land \piv) = \{\w_1,\w_2,\w_3\}$ has probability $\tfrac{9\gamma}{108}$ whereas $t(\state_{\red} \land \piv) = \{w_4\}$ has probability $\tfrac{18\gamma}{108}$, rationalizing a vote for red. This is precisely because the subject does not intuit the true relation between her ability to influence the election and the state.

Now, consider what happens when the computers' votes are themselves observed by the subject. Conditional on seeing split votes, that is, learning that $\vote_{\textsc{s}}$ is true, event-based updating requires that the subject's state-space is $\W^{\vote_{\textsc{s}}} = t(\vote_{\textsc{s}}) = \{w_1, w_5, w_9, w_{13}\}$. If the subject further sees a \red signal, event-based updating indicates $\W^{\vote_{\textsc{s}},\obs_{\red}} = t(\vote_{\textsc{s}}) \cap t(\obs_{\red}) = \{w_1,w_5\}$. Therefore, in every state under consideration, $\state_{\blue}$ is true. 

This example shows how event-based updating alone can make sense of the observed increase in rational behavior in the sequential treatment, where the computers' votes are observed. Nonetheless, we argue that this is not the full story. Notice that if the subject engages in purely event-based updating on the truth of the statement $\vote_{\textsc{s}}$, she remains uncertain as to whether she is pivotal, that is, about the truth of $\piv$. While the experiment itself does not refute this, it seems intuitive that the subject realizes---after seeing $\vote_{\textsc{s}}$ to be true---that she is, indeed, pivotal. That is, by contemplating the computers' voting behavior directly, the subject comes to realize that $\vote_{\textsc{s}} \iff \piv$; as such, her state-space after updating is in fact $\bar{W}^{\vote_{\textsc{s}}} = \{w_1, w_5\}$. This particular hypothesis can be directly tested using our definition of subjective implication: after (but not before) observing the computers votes, the subject should be indifferent between betting on $\{\state_{\blue}, \piv\}$ and $\{\state_{\blue}\}$.
\hfill $\blacksquare$
\end{example}

\section{Interpretation-Dependent Preferences}
\label{sec:SIDEU}

 When the modeler and the \dm entertain different interpretations of uncertainty, descriptions of complex choice objects may not have a unique translation in the mind of the \dm. When the modeler entertains an IOU $(W^*, t^*)$, he views an act as determining payoffs in each state, $\f^*: W^* \to \R$, as in standard  decision theory. However, as the state-space $W^*$ is subjective, he must describe this act to the \dm using their shared language, $\L$. A \emph{syntactic act} is a function from a subset of contingencies into utils; that is, $f: K \to \R$, where $K \subset \K$.\footnote{The set of bets, defined in Section \ref{sec:prelim}, are contained in the set of syntactic acts, and moreover, each act can be thought of as a collection of bets specifying potentially different payoffs.} A syntactic act $f$ \emph{describes} $\f^*$ if for all $\Phi$ in the domain of $f$
 $$
f(\Phi) =  \f^*(w) \text{ for all }  w \in t^*(\Phi).
 $$
 That is, the payoff specified by $f$ contingent on $\Phi$ must coincide with the payoff determined by $\f^*$ in any state where $\Phi$ is true.\footnote{We do  require $\L$ is closed under logical connectives, and so do not require any rationality of the modeler. Nonetheless, given a structured $\L$, it is without loss of generality to consider a rational modeler and so \emph{objectively unambiguous} syntactic acts, those such that $f(\phi) \neq f(\psi)$ implies $t(\phi) \cap t(\psi) = \emptyset$ for all sound $t$. It is also possible to allow modeler himself to be an imperfect logical reasoner.}

Now consider a \dm with IOU $(\W,t)$ and probabilistic beliefs $\mu \in \Delta(W)$ and who considers the syntactic act $f$. Before computing the expected value of $f$, she must first translates $f$ into her model as a function $\f: \W \to \R$. When the modeler and \dm entertain different IOUs, this translation step is not trivial. The act $f$ may not have a unique translation into the state-space of the \dm (i.e., as a function $\W \to \R$). For example, the act $\actG$ shown in Figure \ref{fig:actB}.

Say that $\f: \W \to \R_+$ is \emph{consistent} with $f$ if for all $\w \in \W$
 $$
 \f(w) = f(\Phi) \text{ for some } \Phi \text{ in the domain of $f$ such that } w \in t(\Phi).
 $$
 Notice the difference between the definition of `describes' and `consistent.' In the former, the payoff is uniquely determined (guaranteed by the \emph{for all} quantifier) whereas in the latter it is not (permitted by the \emph{for some} quantifier). Fixing the interpretation of uncertainty let $\cons{f}$ collect the set of all semantic acts consistent with $f$; when multiple statements that yield different payoffs are simultaneously true in some state (of the \dm's state-space), consistency does not pin down the outcome of $f$; $\cons{f}$ is non-singleton. 
Under the interpretation $(\W,t)$, the verbal description of an act, $f$, does not distinguish between the elements of $\cons{f}$. Thus, the \dm's inability to properly interpret the contingencies on which $f$ is predicated creates ambiguity.  

There may be many syntactic descriptions that, from the modeler's perspective, describe the same act. Notice, if $(\W,t) = (\W^*,t^*)$ then for any $f$ describing $\f^*$, we have $\cons{f} = \{\f^*\}$. So when their interpretations coincide, the choice of description is irrelevant, as it will be interpreted flawlessly. When their interpretations differ, however, the \dm's ability to understand the modeler will depend on the latter's description. This reflects the framing effects that arise in the absence of logical omniscience. For example, although ascending clock auctions and second-price sealed-bid auctions share the same normal form (and are thus, from a particular logically unconstrained vantage, strategically equivalent), laboratory subjects are much less likely to employ dominated strategies in the former \citep{kagel1987information,li2017obviously}.

When the \dm encounters a syntactic act $f$ where $\cons{f}$ is non-singleton, she faces ambiguity that arises from the differences between her own and the modeler's interpretations of statements. In what follows, we assume that she responds to this ambiguity with a principle of caution. This caution arises from some mixture of believing the modeler is incorrect---so that the ambiguity reflects the modeler's inconsistent description---or believing she herself is incorrect in some way she is unable to rectify---so that ambiguity reflects her own inconsistent translation. For example, a \dm may be unable to uniquely determine the tax implications of an investment. This could be because there are incompatible statutes in the tax code or because she is unable to properly parse the bounds of when such statutes apply. In either case, we assume that \dm prefers to minimize her exposure to such indeterminacy. 

We consider \emph{sparse interpretation-dependent expected utility} (SIDEU) preferences, reminiscent of \emph{twofold multiprior} preferences as introduced by \cite{echenique2022twofold} in the domain of purely probabilistic uncertainty. 
With SIDEU preferences, $f$ is (strictly) preferred to $g$ if and only if every consistent translation of $f$ has a higher expected payoff than that of $g$. Formally:

\begin{definition}[SIDEU preferences]
Call $\s$ a \df{SIDEU preference} if there exists some IOU $(\W,t)$ and probability $\mu \in \D(\W)$ such that for all syntactic acts $f$ and $g$,
\begin{equation}
\label{eq:sIDEU}
\tag{\textsc{sideu}}
f \succ g \quad \iff  \quad \min_{\f \in \cons{f}}\int\f  \,\textup{d}\l \ > \ \max_{\g \in \cons{g}}\int \g \, \textup{d}\l
\end{equation}

\end{definition}

The SIDEU representation is an interval representation \`{a} la \cite{fishburn1970intransitive}: each act $f$ is associated with an interval of utilities and $f \succ g$ when the $f$'s interval lies entirely above $g$'s. The interval of utilities, is, of course, generated by the set of consistent translations---it represents the range of possible expected payoffs taking into account the ambiguity arising from the \dm's inability to precisely interpret acts. SIDEU reduces to subjective expected utility when the modeler and \dm share an IOU. We provide an axiomatization for sparse IDEU preferences in 
our \href{https://evanpiermont.github.io/pdfs/FCT-SUPP.pdf}{supplemental material}.


\begin{example}
\label{ex:auctions}

\begin{figure}[]
  \centering
  \begin{subfigure}[t]{0.45\textwidth}{}
    \centering
    \begin{tikzpicture}[
                every node/.style   ={font=\footnotesize},
        gridlines/.style    ={gray,dotted,thin},
        x=1.3cm,
        y=0.3cm
    ]
      \def\n{4}
      \def\m{4}

      \foreach \x in {-1,...,\numexpr\n-1}{
          \draw[gridlines] (\x+1,-1) -- (\x+1,\m+.5);
      }
      \foreach \y in {-1,...,\numexpr\m-1}{
          \draw[gridlines] (0,\y+1) -- (\n+.3,\y+1);
      } 

\path[fill=lam2, postaction={pattern={Lines[angle=45,distance={8pt},line width=4pt]}, pattern color=lam1}]
     (0,3.85) rectangle ++(1,.3);
  \path[fill=lam2, postaction={pattern={Lines[angle=45,distance={8pt},line width=4pt]}, pattern color=lam1}]
     (1,2.85) rectangle ++(1,.3);
    \path[fill=lam2] (2,0.85) rectangle ++(1,.3);
     \path[fill=lam2, postaction={pattern={Lines[angle=45,distance={8pt},line width=4pt]}, pattern color=lam1}]
     (3,0.85) rectangle ++(1,.3);

      \path[fill=lam1] (2,1.85) rectangle ++(1,.3);

      \draw[very thick] (0,-1.25) -- (0,\m+0.5);
      \draw[very thick] (-0.25,0) -- (\n+0.25,0);

      \foreach \i in {1,...,\numexpr\n}{
           \node[above=4pt] at (\i-1+0.5,-2.1) {$w_{\number\numexpr\i}$};
      }

      \foreach \i in {1,...,4}{
           \node[above=4pt] at (-.2,\i-1.2) {\scriptsize $\number\numexpr\i-1$};
      }

    \end{tikzpicture}
    \caption{Bids $\actThree$ and $\actFour$ to modeler}
    \label{fig:auctionA}
  \end{subfigure}
  \hfill
  \begin{subfigure}[t]{0.45\textwidth}{}
    \centering
    \begin{tikzpicture}[
                every node/.style   ={font=\footnotesize},
        gridlines/.style    ={gray,dotted,thin},
        x=1.3cm,
        y=0.3cm
    ]
      \def\n{4}
      \def\m{4}

      \foreach \x in {-1,...,\numexpr\n-1}{
          \draw[gridlines] (\x+1,-1) -- (\x+1,\m+.5);
      }
      \foreach \y in {-1,...,\numexpr\m-1}{
          \draw[gridlines] (0,\y+1) -- (\n+.3,\y+1);
      }

      \fill[pattern={Lines[angle=45,distance={1.2pt},line width={0.6pt}]}, pattern color=lam1!20, draw=none] (0,1) rectangle ++(1,3);
      \fill[pattern={Lines[angle=45,distance={1.2pt},line width={0.6pt}]}, pattern color=lam1!20, draw=none] (1,1) rectangle ++(1,2);
      \fill[pattern={Lines[angle=45,distance={1.2pt},line width={0.6pt}]}, pattern color=lam1!20, draw=none] (2,1) rectangle ++(1,1);
      \fill[pattern={Lines[angle=45,distance={1.2pt},line width={0.6pt}]}, pattern color=lam1!20, draw=none] (2,1) rectangle ++(1,1);
      \fill[pattern={Lines[angle=-45,distance={1.2pt},line width={0.4pt}]}, pattern color=lam2!20, draw=none] (0,2) rectangle ++(1,2);
       \fill[pattern={Lines[angle=-45,distance={1.2pt},line width={0.4pt}]}, pattern color=lam2!20, draw=none] (1,2) rectangle ++(1,1);


      \path[fill=lam2, postaction={pattern={Lines[angle=45,distance={8pt},line width=4pt]}, pattern color=lam1}]
     (1,2.85) rectangle ++(1,.3);

     \path[fill=lam2, postaction={pattern={Lines[angle=45,distance={8pt},line width=4pt]}, pattern color=lam1}]
     (0,3.85) rectangle ++(1,.3);

      \path[fill=lam1] (2,1.85) rectangle ++(1,.3);
      \path[fill=lam1] (0,0.85) rectangle ++(2,.3);

      \path[fill=lam2, postaction={pattern={Lines[angle=45,distance={8pt},line width=4pt]}, pattern color=lam1}]
     (2,0.85) rectangle ++(2,.3);

       \path[fill=lam2](0,1.85) rectangle ++(2,.3);


      \draw[very thick] (0,-1.25) -- (0,\m+0.5);
      \draw[very thick] (-0.25,0) -- (\n+0.25,0);

      \foreach \i in {1,...,\numexpr\n}{
           \node[above=4pt] at (\i-1+0.5,-2.1) {$w_{\number\numexpr\i}$};
      }

      \foreach \i in {1,...,4}{
           \node[above=4pt] at (-.2,\i-1.2) {\scriptsize $\number\numexpr\i-1$};
      }

    \end{tikzpicture}
    \caption{Bids $\actThree$ and $\actFour$ according to \dm}
    \label{fig:auctionB}
  \end{subfigure}

  \caption{Discretized version of a second-price auction for a bidder with value $v = 4$.  The acts $\actThree$ and $\actFour$ correspond to a bid of three and four, respectively.}
  \label{fig:auction-pair}
\end{figure}

Consider a bidder in a second-price auction for an object she values at $v > 0$. For each $r \in [0,\bar{r}]$ (with $v < \bar{r})$, consider the (algebra generated by the) statements:
\begin{itemize}[leftmargin=2cm, labelsep=.2em, itemsep=-1.2ex]
\small
  \item[$\phi_r =$] ``A bid $r$ or greater will win the auction''
  \item[$\psi_p = $] ``The winner will pay $p$''
\end{itemize}
The modeler's IOU is $(W,t)$ where $\W = \{\w_q\}_{q \in [0,\bar{r}]}$ where the index $q$ represents the second highest bid. The modeler's truth assignment is such that  $t(\phi_r) = \{\w_q \mid r > q\}$, $t(\psi_p) = \{\w_q \mid p=q\}$ and is otherwise sound; in particular: $t(\phi_r \land \psi_{p}) = t(\phi_r ) \cap t(\psi_{p})$.

Now, consider a bidder who does not fully understand the relation between the winning condition (i.e., $\phi$) and the subsequent payment (i.e., $\psi$). Specifically let $t'$ coincide with $t$ on the atomic statements, but $t(\phi_r\land \psi_p)= \{\w_q \mid q \leq p < r\}$, so that 
she never rules out any payment that is sufficiently large to win the auction.

A syntactic act representing placing a bid $b$ is given by $f_b: \phi_r \land \psi_{p} \mapsto v - p$ for any $r \leq b$ and $0$ otherwise.
From the modeler's perspective, bidding her valuation---the act $f_v$---state-wise dominates any other bid, see Figure \ref{fig:auctionA}.

After translating $f_b$ into her own IOU, $\f_b$ determines the states in which she wins the auction, $\{ \w_q \in \W \mid b > q\}$, but only loosely determines the transfer conditional on winning, $\{p \in \R \mid q \leq p \leq b\}$. As such she cannot rule out the possibility that by lowering her bid she will also lower the transfer. See Figure \ref{fig:auctionB}. For this bidder, $f_v$ is no longer dominant: the worst-case evaluation in $\cons{f_v}$ has a utility of $0$ in all states, whereas the best-case evaluation in $\cons{f_b}$, for $b < v$, has a utility of $v - q$ for states $\w_q$ with $q \leq b$; $f_v$ will not be preferred to $f_b$ according to \eqref{eq:sIDEU}. 

Notice that from the modeler's perspective, $\phi_r \land \psi_p \implies \neg(\phi_{r'} \land \psi_{p'})$ for any $p' < p$; that is, if there is a winning bid resulting in a transfer of $p$, there is no other bid that would reduce this transfer. That captures exactly the intuition behind the incentive compatibility of the second-price auction: the winner's bid determines whether she will win, but not what she will pay. The bidder, however, does not perceive this implication. So by using our definition of subjective implication, we find a direct, testable hypothesis for this explanation of overbidding. 
\hfill$\blacksquare$
\end{example}

\section{Relation to the Literature}
\label{sec:lit}

There is mounting experimental evidence of marked and systematic deviations from rational behavior that extends far beyond what is attributable to subjects' attitudes toward probabilistic uncertainty. For example, behavior that is dependent on the description of the problem \citep{tversky1981framing} or that directly contradicts elementary logical proficiency \citep{tversky1983extensional,walkerjones2023nonBayesian}, the employment of dominated strategies \citep{kagel1993independent,agranov2020non}, extreme reactions to uncertainty \citep{jabarian2022two,kuzmics2022ellsberg,martinez2019failures}, failures of monotonicity \citep{schneider2019experimental}, failures to extract or use available information \citep{enke2020you,araujo2021times}; see \cite{niederle2023failures} for an overview. Indeed, many of these authors directly point to failures of contingent thinking as the driving factor in their findings,
suggesting a strong demand for tractable and universal models of contingent reasoning, and yet---in contrast to the advances in modeling errors in probabilistic judgments---general models of decision-making in the absence of \emph{logical} sophistication are few and far between.

 A notable exception is \cite{lipman-99}, which presents a model in which \dms may consider impossible states of affairs---for example where $p$ and $\neg p$ are both true---called \emph{impossible possible worlds} in the philosophy literature \citep{hintikka1979impossible}. Like us, \cite{lipman-99} constructs an endogenous state-space, the inconsistent states of which account for the \dm's flawed or limited reasoning, but, he focuses directly on the problem of recognizing logical equivalence, without analyzing more general forms of flawed reasoning such as the recognition of implication. His methodology is substantially different from ours; in particular, the objects of choice are acts over an objective/consistent state-space. By contrast, we consider syntactic acts, thereby circumventing the existence of a distinguished `objective' point of view. 

The relation between contingent and probabilistic reasoning is also explored by  \cite{mukerji1997understanding}, \cite{esponda2024contingent}, and \cite{piermont2021hypothetical}. Like above, the choice objects in these models relate to a state-space that is exogenously imposed, limiting the type of inferences that can be drawn. 
SIDEU is closely related to \emph{twofold multiprior preferences}, introduced by \cite{echenique2022twofold} in the domain of risk, themselves motivated by obvious dominance \citep{li2017obviously}.  \citeauthor{echenique2022twofold}~state their representation captures \dms ``who may not be able to fully reason in terms of the underlying state-space," but, because they require an objective state-space to define acts, failures of reasoning appear only implicitly through their effect on probabilistic trade-offs.

Closely related to the methodology of our paper are models of syntactic decision theory, for example \cite{,piermont2017introspective,blume2021constructive,bjorndahl2021language,piermont2023vague} and \cite{piermont2025do}.
Methodologically, the closest paper to ours is \cite{blume2021constructive} (BEH), which also takes as primitive a preference relation over syntactic objects and constructs a state-space as a representation object. Their choice objects are  conditional assignments of the form ``$\textbf{\texttt{if }} \phi \textbf{\texttt{ then }} a_1 \textbf{\texttt{ else }} a_2$,'' where $a_1$ and $a_2$ can be either other such choice objects or primitive (i.e., unconditional) outcomes. Their construction relies critically on the additive structure of their representation and they do not apply their methodology to the study of contingent reasoning.

Finally, the identification of the \dm's subjective model in our paper relates closely to the emerging literature on misspecified models 
\citep{eliaz2020cheating, frick2020misinterpreting,ellis2021subjective}. In contrast to this paper, this literature assumes that \dms, despite having an incorrect view of the world, are logically consistent. 

\appendix

\titlespacing*{\subsection}{0pt}{0pt}{0pt}

\setlength{\abovedisplayskip}{3pt}
\setlength{\belowdisplayskip}{3pt}

\section{Proofs}
\label{sec:proofs}

\def\RR{\mathbf{R}}
\def\f{\mathfrak{f}}
\def\i{\mathfrak{f}^*}
\newcommand{\ext}[1]{[\![#1]\!]}
\newcommand{\extt}[1]{\langle\!\langle#1\rangle\!\rangle}
\newcommand{\meet}[2]{#1\odot#2}
\newcommand{\cover}[2]{#1\oplus#2}

\subsection{Preliminaries: Taxonomy of Filters}
For any partial order, $\RR$, over a set $X$, generalize the language from the body of the paper: call $x \in X$ \emph{null} if $x\RR y$ for all $y \in X$ and call $x,y$ \emph{disjoint}, written $x \bot y$, if there exists no nonnull $z$ such that $z \RR x$ and $z \RR y$. A subset $Y\subseteq X$ is a \emph{$\RR$-up-set} (or, simply \emph{up-set} when the relation is obvious) if whenever $x \in Y$ and $x\RR y$ then $y \in Y$. A nonempty up-set $Y$ is called a \emph{$\RR$-filter}, (or \emph{filter}) whenever $x,y \in Y$ then there exists a $z \in Y$ such that $z \RR x$ and $z \RR y$. 
A filter is called \emph{proper} if it does not contain any null element.\footnote{Our definition of \emph{proper} differs slightly from the standard usage, where a proper filter is one that is not the totality of $X$. Should there exist any null element, these two definitions coincide. However, when there are no null elements (for example $\mathbb{Z}$ under $\leq$), then we allow the full set $X$ to be proper.}
Let $\mathfrak{F}(\RR,X)$ collect all proper filters over $\RR$. It is easy to verify that $\mathfrak{F}(\RR,X)$ is closed under intersections. A proper filter is called an \emph{ultra}filter if it cannot be extended (by set inclusion) to any other proper filter. Let $\mathfrak{U}(\RR)$ collect all ultrafilters over $\RR$.  

We say that a proper filter $\mathfrak{f} \in  \mathfrak{F}(\RR,X)$ is \textit{irreducible}, if, for every $\mathfrak{g}_{1},\mathfrak{g}_{2}\in \mathfrak{F}(\RR,X)$,
\[
\mathfrak{g}_{1}\cap\mathfrak{g}_{2}=\mathfrak{f} \text{ implies } \mathfrak{g}_{1}\subseteq\mathfrak{f}\hspace{0.2cm}\textup{or}\hspace{0.2cm}\mathfrak{g}_{2}\subseteq\mathfrak{f}.
\]
Similarly, we say that $\mathfrak{f}$ is \textit{prime}, if, for every $\mathfrak{g}_{1},\mathfrak{g}_{2}\in  \mathfrak{F}(\RR,X)$,
\[
\mathfrak{g}_{1}\cap\mathfrak{g}_{2}\subseteq\mathfrak{f}\hspace{0.2cm}\textup{implies}\hspace{0.2cm}\mathfrak{g}_{1}\subseteq\mathfrak{f}\hspace{0.2cm}\textup{or}\hspace{0.2cm}\mathfrak{g}_{2}\subseteq\mathfrak{f}.
\]
Let $\mathfrak{I}(\RR,X)$ and $\mathfrak{P}(\RR,X)$ collect all irreducible and prime filters, respectively,  of $\RR$ on $X$. 

 Say that $\RR$ \emph{has meets} if for all $x,y \in X$ such that there exists a $z$ such that $z \RR x$ and $z \RR y$, then there exists an element $\meet{x}{y} \in X$, referred to as the meet of $x$ and $y$, such $z \RR x$ and $z \RR y$ if and only if $z \RR(\meet{x}{y})$. The following are immediate to derive: $(\meet{x}{y}) = (\meet{y}{x})$, $(\meet{x}{x}) = x$,  $(\meet{x}{y})\RR x$, and if $y\RR y'$ then $(\meet{x}{y})\RR(\meet{x}{y'})$. 

 Say that $\RR$ \emph{has covers} if for all $x,y \in X$ there exists an element $\cover{x}{y} \in X$, referred to as a \textit{cover} of $x$ and $y$, such that $x \RR (\cover{x}{y})$, $y \RR (\cover{x}{y})$ and such that $z\bot x$ and $z \bot y$ if and only if $z \bot (\cover{x}{y})$.

\begin{lemma}
\label{lem:meetsfilter}
Let $\RR$ be a partial order over a set $X$ that has meets. Let $\f \in \mathfrak{F}$ and $z \in X$ be such that $z$ is not disjoint from any $x\in \f$. Then
$
\f_z = \{y \in X \mid (x\odot z)\RR y, x \in \f\} 
$
is a proper filter containing $\f$. 
\end{lemma}

\begin{proof}
Consider $y \in \f_z$ and $y\RR y'$. Then  there is some $x \in \f$ such that $(x\odot z)\RR y \RR y'$; $\f_z$ is an up-set. Now let $y,y' \in \f_z$ so that there is some $x,x' \in \f$ such that $(x\odot z)\RR y$ and  $(x' \odot z)\RR y'$. Since $\f$ is a filter, there exists some $x'' \in \f$ such that $x''\RR x$ and $x''\RR x'$. By the properties of meets, $(x'' \odot z)\RR(x\odot z) \RR y$ and $(x'' \odot z)\RR(x' \odot z) \RR y'$. Since $(x \odot z)\RR x$, we have that $\f \subseteq \f_z$. 
\end{proof}

\begin{lemma}
\label{lem:sepoints}
Let $\RR$ be a partial order over a set $X$ and let $\f \in \mathfrak{F}$ and $x \in X$ be such that $x \notin \f$. Then there exists an irreducible filter, $\i \in \mathfrak{I}(\RR,X)$, such that $\f \subseteq \f^*$ and $x \notin \i$. Moreover, if $\RR$ has meets and covers, and if $x\bot y_0$ for some $y_0 \in \f$, then it can be assumed that $\i$ is prime: $\i \in \mathfrak{P}(\RR,X)$.
\end{lemma}

\begin{proof}
Consider $F  = \{ \f' \in  \mathfrak{F} \mid \f \subseteq \f', x \notin \f'\}$, ordered by set-inclusion. Let $C$ be a chain of $F$ and let $\mathfrak{c} = \bigcup C$. 
That $\mathfrak{c} \in F$, hence $\mathfrak{c}$ is an upper-bound for $C$, follows standard arguments. 
We can thus apply Zorn's lemma to find a maximal element, $\i$, of $F$. We must show that $\i$ is irreducible. Since $\i$ is maximal in $F$, it follows that, for any $\f' \in \mathfrak{F}(\RR)$ if $\i \subsetneq \f'$ then $x \in \f'$. Thus $x \in \bigcap \{\f' \in \mathfrak{F}(\RR), \i \subsetneq \f' \}$ but $x \notin \i$; $\i$ is irreducible. 

Now, assume in addition that $\RR$ has meets and covers and that $x\bot y_0$ for some $y_0 \in \f$. First, we claim that for each $z \notin \f^*$, $z$ is disjoint from some element of $\i$. Assume the contrary, so we have by Lemma \ref{lem:meetsfilter} that $\i_z = \{y' \in X \mid (y \odot z)\RR y', y \in \i\}$ is a filter containing $\f^*$; by our assumption of maximality, $x \in \i_z$ (if it did not contain $x$, it would be proper filter extending $\i$). So then: $(y \odot z) \RR x$ for some $y \in \i$. Since $\i$ is a filter, there exists some $y' \in \i$ such that $y' \RR y$ and $y' \RR y_0$; moreover, $(y' \odot z)$ is not null since $y'$ is not disjoint from $z$, by assumption. Finally, we have  $(y' \odot z)\RR (y \odot z) \RR x$ and $(y' \odot z)\RR y' \RR y_0$, contradicting the assumption that $y_0$ was disjoint from $x$.

We now show that $\i$ is prime. To see this: pick two filters $\mathfrak{g}_{1},\mathfrak{g}_{2}\in \mathfrak{F}(\RR,X)$ where $\mathfrak{g}_{1}\cap\mathfrak{g}_{2}\subseteq\i$. Then, proceed by contradiction and suppose that $\mathfrak{g}_{1}\nsubseteq\i$ and $\mathfrak{g}_{2}\nsubseteq\i$. We can then pick $z_1 \in\mathfrak{g}_{1} - \i$ and  $z_2 \in\mathfrak{g}_{2} - \i$ (notice that $z_1 \neq z_2$). Then, as seen in the prior claim, there exist $y_1,y_2 \in\i$ such that $y_1\bot z_1$ and $y_2 \bot z_2$. Since $\i$ is a filter, there is some $y' \in \i$ where $y' \RR y_1$ and $y' \RR y_2$ and notice, then, $y' \bot z_1$ and $y' \bot z_2$. There exists some cover of $z_1$ and $z_2$, $(z_1 \oplus z_2)$, and we have $y' \bot (z_1 \oplus z_2)$. Moreover, since $z_1 \RR (z_1 \oplus z_2)$ and $z_2 \RR (z_1 \oplus z_2)$, we have $(z_1 \oplus z_2) \in \mathfrak{g}_{1}\cap\mathfrak{g}_{2}\subseteq\i$, a contradiction, since it is disjoint from $y' \in \i$.
\end{proof}

\begin{lemma}
\label{lem:irr*}
If $\f\in  \mathfrak{I}(\RR,X)$ is an irreducible filter, then for $\mathfrak{g}_{1} \ldots \mathfrak{g}_{n}\in \mathfrak{F}(\RR,X)$, with $n \geq 2$,
$\bigcap_{i\leq n} \mathfrak{g}_i =\mathfrak{f}$ implies $\mathfrak{g}_{i} = \f$ for some $ i \leq n$.
\end{lemma}

\begin{proof}
We prove this by induction on $n$. The base case, for $n=2$ is due to irreducibility. Assume this holds for all intersections of $n-1$ and fix some $\mathfrak{g}_{1} \ldots \mathfrak{g}_{n}\in \mathfrak{F}(\RR,X)$ such that $\bigcap_{i\leq n} \mathfrak{g}_i =\mathfrak{f}$. Clearly, for each $i \leq n$, $\f \subseteq \mathfrak{g}_i$. Let $\mathfrak{g} = \bigcap_{2 \leq i\leq n} \mathfrak{g}_i$; since the intersections of filters is a filter, the fact that $\mathfrak{g}_1 \cap \mathfrak{g} = \f$ implies via irreducibility that either $\mathfrak{g}_{1} \subseteq \f$ or $\mathfrak{g} \subseteq \f$. By the inductive hypothesis, we have the result.
 \end{proof}

 \subsection{Properties of Implication and Interpretation}

\begin{lemma}
\label{lem:weakorder}
If $\s$ is reflexive and transitive then $\imp$ is reflexive and transitive.
\end{lemma}

\begin{proof}
 We must show that $\imp$ is transitive (reflexivity is immediate): so let $\Phi \imp \Psi$ and $\Psi \imp \Lambda$. Thus we have, for $x > 0$, and $\Gamma \in \K$:
\begin{align*}
x_{\Lambda \cup \Gamma} \sim x_{\Lambda \cup \Psi \cup \Gamma} \sim x_{\Lambda \cup \Psi \cup \Phi \cup \Gamma} \sim x_{\Lambda \cup \Phi\cup \Gamma} 
\end{align*}
and so $\Phi \imp \Lambda$, as needed.
 \end{proof}

 \begin{lemma}
\label{lem:subsetorder}
Let $\s$ be a reflexive and transitive, then the following hold
\begin{enumerate}[label=\textup{(\roman*)}, itemsep=-.5ex]
  \item \label{sso:welldef} $\Phi \imp \Psi$ if and only if $\Phi \cup \Psi \imp \Psi$. 
  \item \label{sso:subset}  $\Phi \subseteq \Psi$, then $\Phi \imp \Psi$,
  \item \label{sso:disjoint}  $\Phi \bot \Psi$ and $\Phi' \imp \Phi$ implies $\Phi' \bot \Psi$.
  \item \label{sso:union}  $\Phi \imp \Psi$ and $\Psi \imp \Psi$ if and only if $\Phi \cup \Phi' \imp \Psi$.
\end{enumerate}
\end{lemma}

\begin{proof}
\ref{sso:welldef} follows immediately from the fact that, for all $x \in \R$ and $\Gamma \in \K$: $x_{\Psi \cup \Phi \cup \Gamma} = x_{\Psi \cup (\Phi \cup \Psi)\cup \Gamma}$. 
\ref{sso:subset} follows from reflexivity of $\imp$ and part \ref{sso:welldef}.
To see \ref{sso:disjoint}, let $\Gamma \imp \Psi$, and $\Gamma \imp \Phi'$, then by the transitivity of $\imp$,  $\Gamma \imp \Phi$, so by the disjointness of $\Phi$ and $\Psi$, $\Gamma$ must be null.

The `if' direction of \ref{sso:union} follows from \ref{sso:subset} and transitivity. Towards the only if, let  $\Phi \imp \Psi$ and $\Phi' \imp \Psi$. Then we have by definition of $\imp$, that for all $x \in \R$ and $\Gamma \in \K$, that $x_{\Psi \cup \Gamma} \sim x_{\Psi \cup \Phi' \cup \Gamma}$ and $x_{\Psi \cup (\Phi' \cup \Gamma)} \sim x_{\Psi \cup \Phi \cup (\Phi' \cup \Gamma)}$. Thus, by transitivity, we have $x_{\Psi \cup \Gamma} \sim x_{\Psi \cup (\Phi \cup \Phi') \cup \Gamma}$, or that $\Phi \cup \Phi' \imp \Psi$.
 \end{proof}

As $\imp$ is a pre-order (Lemma \ref{lem:weakorder}), subjective equivalence is a congruence relation. Let
$$\ext{\Phi} = \{ \Psi \in \K \mid \Psi \imp \Phi \text{ and } \Phi \imp \Psi \} $$
denote the equivalence class containing $\Phi$.  In an abuse of notation, we write $\ext{\phi}$ rather than $\ext{\{\phi\}}$.
Let $\K^\dag  = \{\ext{\Phi} \subseteq \K \mid \Phi \in \K \}$ denote the quotient of $\K$ under subjective equivalence. 
$\K^\dag$ inherits the pre-order, $\imp$, as $\ext{\Phi} \imp \ext{\Psi}$ iff $\Phi \imp \Psi$; over $\K^\dag$, $\imp$ is a partial order.
For each $\Phi \in \K^\dag$, denote $\f_\Phi = \{\ext{\Psi} \in \K^\dag \mid  \Phi \imp \Psi \}$. It is trivial to show that $\f_\Phi$ is a filter and it is proper iff $\Phi$ is not null.

 \begin{lemma}
\label{lem:hasmeets}
If $\s$ is articulate, then the partial order $\imp$ over $\K^\dag$ has meets and covers.
\end{lemma}

\begin{proof}
Let $\ext{\Phi},\ext{\Phi'} \in \K^\dag$ be such that there exists some $\ext{\Psi} \in \K^\dag$ such that $\ext{\Psi}  \imp \ext{\Phi} $ and $\ext{\Psi} \imp \ext{\Phi'}$. Now, consider $\Psi^* = \{\psi \in \L \mid \psi \imp \Phi \text{ and } \psi \imp \Phi'\}$, which is nonempty by our assumptions and Lemma \ref{lem:subsetorder}.

We claim that $\ext{\Psi^*}$ is a meet for $\ext{\Phi},\ext{\Phi'}$. To see it, consider $\ext{\Psi} \in \K^\dag$ such that $\ext{\Psi}  \imp \ext{\Phi} $ and $\ext{\Psi} \imp \ext{\Phi'}$. Then, by Lemma \ref{lem:subsetorder}\ref{sso:subset}, we have $\ext{\psi} \imp \ext{\Phi}$ and $\ext{\psi} \imp \ext{\Phi'}$ for all $\psi \in \Psi$, and thus, $\Psi \subseteq \Psi^*$. It follows from yet one more application of Lemma \ref{lem:subsetorder}\ref{sso:subset} that $\ext{\Psi} \imp \ext{\Psi^*}$.

It remains to be shown that $\ext{\Psi^*}  \imp \ext{\Phi}$ and $\ext{\Psi^*} \imp \ext{\Phi'}$. 
Let $\Gamma \bot \Phi$, then we have that $\Gamma \bot \psi$ for all $\psi \in \Psi^*$ by  Lemma \ref{lem:subsetorder}\ref{sso:disjoint}. Thus, by part \ref{art2}, we have $\Gamma \bot \Psi^*$. Thus, there is no nonnull $\Gamma$ such that $\Gamma \imp \Psi^*$ and $\Gamma \bot \Phi$. By (the contrapositive of) part \ref{art1}, $\Psi^* \imp \Phi$, as needed. An analogous argument shows $\Psi^* \imp \Phi'$ as well.

It is trivial to check that for any $\Phi, \Psi \in \K$, part \ref{art2} (and Lemma \ref{lem:subsetorder}) implies that $\ext{\Phi \cup \Psi}$ is a cover for $\ext{\Phi}$ and $\ext{\Psi}$.
\end{proof}

\begin{definition}[Canonical interpretation]
\label{def:canonical}
Let $\s$ be a reflexive and transitive preference over syntactic acts. Then, the \df{canonical interpretation} of $\s$ is the pair $(\W^{*},t^{*})$, where:
\[
\W^{*}:=\left\{
\begin{tabular}{l l}
$\mathfrak{I}(\imp, \K^\dag)$&$\textup{if $\s$ is not articulate}$,\\
$\mathfrak{P}(\imp, \K^\dag)$&$\textup{if $\s$ is articulate}$,
\end{tabular}
\right.
\]
and $t^{*}:\K \to 2^{\W^{*}}$, where $t^{*}(\Phi):=\{\w\in\W^{*}\,|\,\ext{\Phi}\in\w\}$ for every $\Phi\in\K$.
\end{definition}

\begin{tproof}{thm:iounoprob}
We will now show that the three properties of faithfulness hold for the \hyperref[def:canonical]{canonical interpretation}:

\begin{enumerate}[leftmargin=0cm]
\item[\ref{faithful:subset}]  Consider $\Phi \imp \Psi$ and $\w \in t^*(\Phi)$. Then, by construction, $\ext{\Phi} \in \w$, and since $\w$ is a $\imp$-up-set, $\ext{\Psi} \in \w$ and, therefore, $\w \in t^*(\Psi)$, so $t^*(\Phi) \subseteq t^*(\Psi)$. 

Now, assume that $\Phi \centernot{\imp} \Psi$. Clearly $\ext{\Phi} \in \f_\Phi$ and $\ext{\Psi} \notin \f_\Phi$, and so, by Lemma \ref{lem:sepoints}, we have the existence of some irreducible filter $\w \in \mathfrak{I}(\imp, \K^\dag)$ such that  $\ext{\Phi} \in \w$ and $\ext{\Psi} \notin \w$. Then, if $\s$ is not articulate, $\w \in t^*(\Phi)$ and $\w \notin t^*(\Psi)$, or, $t^*(\Psi) \not\subseteq t^*(\Phi)$. 

If in addition $\s$ is articulate, then $\Phi \centernot{\imp} \Psi$ implies the existence of some $\Gamma$ such that $\Gamma \imp \Phi$ and $\Gamma \bot \Psi$. Consider $\f_\Gamma \supseteq \f_\Phi$. By Lemma \ref{lem:hasmeets}, $\imp$ (over $\K^\dag$) has meets and covers, and $\Gamma \in \f_\Gamma$ is disjoint from $\Psi$, so by Lemma \ref{lem:sepoints}
we have the existence of some prime filter $\w' \in \mathfrak{P}(\imp, \K^\dag)$ such that  $\ext{\Phi} \in \w'$ and $\ext{\Psi} \notin \w'$. Then, if $\s$ is articulate, $\w' \in t^*(\Phi)$ and $\w' \notin t^*(\Psi)$, or, $t^*(\Psi) \not\subseteq t^*(\Phi)$.

\item[\ref{faithful:null}]   If $\Phi$ is null then $\f_\Phi = \K^\dag$. As such, no proper (hence, irreducible or prime) filter can contain $\ext{\Phi}$: $t^*(\Phi) = \emptyset$.

Now assume that $\Phi$ is not null. Then there exists some $\Psi$ such that $\Phi \centernot{\imp} \Psi$. By part (i), $t^*(\Phi) \not\subseteq t^*(\Psi)$ and thus, $t^*(\Phi) \neq \emptyset$.

\item[\ref{faithful:disjoint}]  Now assume $\w \in t^*(\Phi) \cap t^*(\Psi) \neq \emptyset$. By construction, $\w$ is a proper filter that contains $\ext{\Phi}$ and $\ext{\Psi}$. By virtue of being a filter, $\w$ contains some $\ext{\Gamma}$ such that $\Gamma \imp \Phi$ and $\Gamma \imp \Psi$, and by construction, $\w \in t^*(\Gamma)$. \hfill \qedhere

\end{enumerate}
\end{tproof}

 \begin{lemma}
\label{lem:statesarefilters}
If $(\W,t)$ is a faithful interpretation of $\s$, then for all $\w \in \W$,   
\begin{equation}
\label{eq:thingithinkiscalledtheextension}
\extt{w} = 
\{\ext{\Phi} \in \K^\dag \mid \w \in t(\Phi) \}
\end{equation}
is either empty or a proper $\imp$-filter.
\end{lemma}

\begin{proof}
Let $\w \in \W$ such that $\extt{\w} \neq \emptyset$.
Consider $\ext{\Phi} \in \extt{\w}$ and $\Phi \imp \Psi$. Then $\w \in t(\Phi) \subseteq t(\Psi)$, by\ref{faithful:subset}. so $\ext{\Psi} \in \extt{\w}$; $\ext{\Phi}$ is a $\imp$-upset.
Now let $\ext{\Phi}, \ext{\Psi} \in \extt{\w}$. So by \ref{faithful:disjoint}, we have that $\w \in t(\Gamma)$ for some $\Gamma$ such that $\Gamma \imp \Phi$ and $\Gamma \imp \Psi$. Thus, $\ext{\Gamma} \in \extt{\w}$, and  $\ext{\Gamma} \imp \ext{\Phi}$ and $\ext{\Gamma} \imp \ext{\Psi}$. That is, $\extt{\w}$ is proper then follows immediately from \ref{faithful:null}.
\end{proof}

 \begin{lemma}
\label{lem:meetsintersection}
If $(\W,t)$ is a faithful interpretation of $\s$, and $\Gamma$ is such that $\ext{\Gamma}$ is the meet of $\ext{\Phi}$ and $\ext{\Psi}$, then $t(\Gamma) = t(\Phi) \cap t(\Psi)$.
\end{lemma}

\begin{proof}
By definition of a meet, we have $\Gamma \imp \Phi$ and $\Gamma \imp \Psi$. It follows from \ref{faithful:subset} that $t(\Gamma) \subseteq t(\Phi) \cap t(\Psi)$.
Further, let $w \in t(\Phi) \cap t(\Psi)$; by \ref{faithful:disjoint}, $\w \in t(\Gamma')$ for some $\Gamma'$ such that $\Gamma' \imp \Phi$ and $\Gamma' \imp \Psi$. Thus, by definition of a meet, $\Gamma' \imp \Gamma$, so by \ref{faithful:subset}, $\w \in t(\Gamma') \subseteq t(\Gamma)$. It follows that $t(\Phi) \cap t(\Psi) \subseteq t(\Gamma)$.
\end{proof}

\begin{tproof}{thm:maxstatespace} (1) Fix $\s$. For the maximal state-space define $(\W^+, t^+)$ as 
\[
\W^{+}:= \mathfrak{F}(\imp, \K^\dag) \cup \{\emptyset\},
\]
and $t^{+}:\K \to 2^{\W^{+}}$, as $t^{+}(\Phi):=\{\w\in\W^{+}\,|\,\ext{\Phi}\in\w\}$ for every $\Phi\in\K$. It is a routine translation of the proof of Theorem \ref{thm:iounoprob} to verify that this is indeed a faithful interpretation of $\s$. Given this, the maximality of $(\W^+, t^+)$ follows immediately from Lemma \ref{lem:statesarefilters}.

(2) Assume that $\K^\dag$ is finite and consider the \hyperref[def:canonical]{canonical interpretation} $(\W^*, t^*)$.
 To see that $(\W^*, t^*)$ is minimal, let $(\W,t)$ denote any faithful interpretation of $\s$ and $\f \in \W^*$ be an irreducible filter. We will show that $\f = \extt{w'}$ for some $\w' \in \W$. Since $\K^\dag$ is finite, $\f$ is of the form $\{\ext{\Psi} \in \K^\dag \mid  \Phi \imp \Psi \}$ for some non-null $\Phi$ (i.e., all filters are principal).

Now, for all $\w \in t(\Phi)$, $\extt{\w}$ is a proper filter containing $\Phi$ (by Lemma \ref{lem:statesarefilters}) and thus $\f \subseteq \extt{\w}$.
Moreover, if $\ext{\Gamma} \in \bigcap_{\w \in t(\Phi)} \extt{\w}$ then $t(\Phi) \subseteq t(\Gamma)$. By faithfulness, therefore, $\Phi \imp \Gamma$, and so $\ext{\Gamma} \in \f$. Together these observations imply $\f = \bigcap_{\w \in t(\Phi)} \extt{\w}.$
Lemma \ref{lem:irr*} delivers that $\f = \extt{\w}$ for some $\w \in t(\Phi)$.



(3) Now, assume that $(\W,t)$ extends some $(\W^-,t^-)$ and is extended by some $(\W^+,t^+)$, both faithful to $\s$. We must show that $(W,t)$ is faithful $\s$. This is routine, so we will show only \ref{faithful:subset}, omitting the others:

Let $\Phi \imp \Psi$. Let $\w \in \W$ be such that $\w \in t(\Phi)$. Then there exists a $w^+ \in \W^+$ with $\extt{w^+} = \extt{\w}$. By \ref{faithful:subset} for $(\W^+,t^+)$, we have $\ext{\Psi} \in \extt{\w^+} = \extt{\w}$. 
We have $\w \in t(\Psi)$, so $t(\Phi) \subseteq t(\Psi)$. 

Let $\Phi \centernot{\imp} \Psi$. By \ref{faithful:subset} for $(\W^-,t^-)$, there exists a $\w^- \in \W^-$ with $\w^- \in t^-(\Phi) - t^-(\Psi)$. We have $\ext{\Phi} \in \extt{\w^-}$ and $\ext{\Psi} \notin \extt{\w^-}$. There exists a $w \in \W$ with $\extt{w} = \extt{\w^-}$ and so $\ext{\Phi} \in \extt{\w}$ and $\ext{\Psi} \notin \extt{\w}$.
We have $\w \in t(\Psi) - t(\Psi)$, and so, $t(\Phi) \centernot{\subseteq} t(\Psi)$. 
\end{tproof}

% ...

\begin{tproof}{thm:rep-all}
Let $(\W,t)$ denote an arbitrary faithful interpretation of $\s$. The proof regarding \refax{E} and \refax{I} is trivial; we show \refax{C}:
\begin{enumerate}[leftmargin=0cm]



  \item[(\refax{C})] Assume that $\s$ satisfies \refax{C}. Then $\ext{\phi \land \psi}$ is a meet for $\ext{\phi}$ and $\ext{\psi}$; that that $t(\phi \land \psi) = t(\phi) \cap t(\psi)$ follows from Lemma \ref{lem:meetsintersection}.
Now assume that that $t(\phi \land \psi) = t(\phi) \cap t(\psi)$ for all $\phi,\psi \in \L$. Let $\Gamma \imp \phi$ and $\Gamma \imp \psi$, and $t(\Gamma) \subseteq t(\phi) \cap t(\psi) = t(\phi \land \psi)$, so $\Gamma \imp \phi \land \psi$. Now, let $\Gamma \imp \phi \land \psi$, or, $t(\Gamma) \subseteq t(\phi \land \psi) =  t(\phi) \cap t(\psi)$, implying $\Gamma \imp \phi$ and $\Gamma \imp \psi$. So $\s$ satisfies \refax{C}. \qedhere
\end{enumerate}
\end{tproof}

\begin{tproof}{thm:rep-art}
We will show that the \hyperref[def:canonical]{canonical interpretation} is set-consistent. 
First notice that Lemma \ref{lem:subsetorder}\ref{sso:union} implies that $\f_{\Phi\cup\Psi} =\f_{\Phi} \cap \f_{\Psi}$. 
Now, for any $\w \in\W^*$, we know that $\w\in t^*(\Phi\cup\Psi)$ if and only if $\Phi\cup\Psi \in\w$, that is, if and only if $\f_{\Phi\cup\Psi}\subseteq \w$. As $\w$ is a prime filter, the latter holds if and only if $\f_\Phi \subseteq\w$, or $\f_\Psi \subseteq\w$, and hence, if and only if $\w \in t^*(\Psi)$ or $\w \in t^*(\Psi)$. Thus, as required $t^*(\Phi\cup\Psi)=t^*(\Phi) \cup t^*(\Psi)$.

Now, let $(\W,t)$ denote an arbitrary set-consistent faithful interpretation of $\s$. 

\begin{enumerate}[leftmargin=0cm]
\item[(\refax{D})] 
Assume that $\s$ satisfies \refax{D}. By the reflexivity of $\s$, we have that $\phi \lor \psi \imp \phi \lor \psi$, and thus, by \refax{D}, $\phi \imp \phi \lor \psi$ and $\psi \imp \phi \lor \psi$. By Lemma \ref{lem:subsetorder}\ref{sso:union}, $\{\phi, \psi\} \imp \phi \lor \psi$. 
Similarly, $\phi \imp \{\phi, \psi\}$ and $\psi \imp \{\phi, \psi\}$ by Lemma \ref{lem:subsetorder}\ref{sso:subset}, and thus, by \refax{D}, $\phi \lor \psi \imp \{\phi, \psi\}$. Together, via \ref{faithful:subset}, we have $t(\phi \lor \psi) = t(\{\phi,\psi\})$. By set-consistency we can then conclude that $t(\phi \lor \psi) = t(\{\phi,\psi\}) = t(\phi) \cup t(\psi)$.

Now assume $t(\phi \lor \psi) = t(\phi) \cup t(\psi)$ for all $\phi,\psi \in \L$. Take some $\Gamma$. Then  $\phi \imp \Gamma$ and $\psi \imp \Gamma$ if and only if $t(\phi) \subseteq t(\Gamma)$ and  $t(\psi) \subseteq t(\Gamma)$, if and only if $t(\phi \lor \psi) = t(\phi) \cup t(\psi) \subseteq t(\Gamma)$, if and only if $\psi \lor \psi \imp \Gamma$. So $\s$ satisfies \refax{D}.

\item[(\refax{N})] 
Assume that $\s$ satisfies \refax{N}: 
Since $\phi \bot \neg\phi$, no proper filter contains both $\phi$ and $\neg\phi$. Since every $\Gamma \imp \{\phi, \neg\phi\}$, every proper filter contains $\{\phi, \neg\phi\}$. Thus by Lemma \ref{lem:statesarefilters}: $t(\phi) \cap t(\neg\phi) = \emptyset$ and $t(\{\phi,\neg\phi\}) = \bigcup_{\Phi \in \K}t(\Phi)$. The latter implies via set-consistency that $t(\phi) \cup t(\neg\phi) = \bigcup_{\Phi \in \K}t(\Phi)$, and so $\{t(\phi),t(\neg\phi)\}$ partitions $\bigcup_{\Phi \in \K}t(\Phi)$.

Now, assume that $t(\neg \phi) = \bigcup_{\Phi \in \K}t(\Phi) - t(\phi)$ for all $\phi \in \L$. Then $t(\phi) \cap t(\neg\phi) = \emptyset$, so by faithfulness, $\phi \bot \neg\phi$. Moreover, $t(\{\phi,\neg\phi\}) = \bigcup_{\Phi \in \K}t(\Phi)$ so $\Gamma \imp \{\phi, \neg\phi\}$ for all $\Gamma \in \K$.  So $\s$ satisfies \refax{N}.\qedhere
\end{enumerate}
\end{tproof}

\begin{tproof}{thm:standardUpdating} Recall that $\mathfrak{F}(\imp, \K^\dag)$ is the set of all $\imp$ filters and $\mathfrak{F}(\impp, \K^\dag)$ is the set of all $\impp$ filters. We will show that the following is equivalent to event-based updating 

\begin{axiom}{}{BS}{Bi-Separation} 
If $\Psi \notimpp \Lambda$ then there exists some $\f \in \mathfrak{F}(\imp, \K^\dag) \cap \mathfrak{F}(\impp, \K^\dag)$ such that $\Psi \in \f$ and $\Lambda \notin \f$. 
\end{axiom}

\begin{lemma}
\label{lem:BS}
$(\s, \ss)$ is event-based if and only if it satisfies \refax{BS}.
\end{lemma}

\begin{subproof}
Towards necessity, assume event-based updating via $(\W,\W^{\circ},t)$. Assume $\Psi \notimpp \Lambda$, then there exists some $\w \in \W$ such that $\w \in t(\Psi) \cap \W^{\circ}$ and $w \notin t(\Lambda)\cap \W^{\circ}$. Thus we have that $\w \in t(\Psi)$ and $\w \notin t(\Lambda)$, or, that $\Psi \in \extt{\w}$ and $\Lambda \notin \extt{\w}$. Finally, Lemma \ref{lem:statesarefilters} ensures us that $\extt{\w} \in \mathfrak{F}(\imp, \K^\dag) \cap \mathfrak{F}(\impp, \K^\dag)$.

Towards sufficiency, assume \refax{BS} holds. Let $\W$ denote the set of all filters of $\imp$, with $t$ the inclusion map. As shown in the proof of Theorem \ref{thm:maxstatespace}, this is a (maximal) faithful interpretation. Let 
$$\W^{\circ} = \{w \in \W \mid w \text{ is a } \impp \text{filter} \}.$$

By virtue of each $\w \in \W^+$ being a $\impp$-filter, it is immediate that $W^{\circ}$ satisfies \ref{faithful:null} and \ref{faithful:disjoint}. We need only show (i): if $\Psi \impp \Lambda$ then, since each $\w \in \W^{\circ}$ is an $\impp$-up-set, we have that $t^{\circ}(\Psi) \subseteq t^{\circ}(\Lambda)$. If instead, $\Psi \notimpp \Lambda$ then by \refax{BS} there exists some $\f \in \mathfrak{F}(\imp, \K^\dag) \cap \mathfrak{F}(\impp, \K^\dag)$ such that $\Psi \in \f$ and $\Lambda \notin \f$. By construction $\f \in \W^{\circ}$, so in particular $\f \in t^{\circ}(\Psi) - t^{\circ}(\Lambda)$, and so $t^{\circ}(\Psi) \centernot{\subseteq} t^{\circ}(\Lambda)$.
\end{subproof}

We will now show that if $(\s, \ss)$ satisfies \refax{BS}, then it satisfies \refax{DC-A} and \refax{DC-B}, and the converse holds so long as the set of $\imp$-equivalence classes is finite. By Lemma \ref{lem:BS}, this proves the theorem. First assume \refax{BS} holds:

(\refax{DC-A}) We will show the contrapositive: Assume $\Psi \notimpp \Lambda$, so by \refax{BS} there exists some $\f \in \mathfrak{F}(\imp, \K^\dag) \cap \mathfrak{F}(\impp, \K^\dag)$ such that $\Psi \in \f$ and $\Lambda \notin \f$.
Since $\f$ is upwards-$\imp$-closed, it follows that $\Psi \centernot{\imp} \Lambda$.

(\refax{DC-B}) Let $\Psi \impp \Gamma$ and $\Psi \impp \Gamma'$ and $\Psi \notimpp \Lambda$. By \refax{BS} there exists an $\f \in \mathfrak{F}(\imp, \K^\dag) \cap \mathfrak{F}(\impp, \K^\dag)$ such that $\Psi \in \f$ and $\Lambda \notin \f$. Since $\f$ is upwards-$\impp$-closed, $\{\Gamma,\Gamma', \Psi\} \subseteq \f$. Since $\f$ is $\imp$-directed, there exists some $\Psi' \in \f$ such that $\Psi' \imp \Psi$, $\Psi' \imp \Gamma$, and $\Psi' \imp \Gamma'$; by \refax{DC-A}, as just established, $\Psi' \impp \Psi$. Finally, since $\f$ is upwards-$\impp$-closed, it follows that $\Psi' \notimpp \Lambda$.

Now, further assume that the set of $\imp$-equivalence classes is finite. Let $\Psi \notimpp \Lambda$. Consider $F  = \{ \f' \in  \mathfrak{F}(\impp, \K^\dag) \mid \Psi \in \f', \Lambda \notin \f'\}$ ordered by set-inclusion, and by standard arguments, we can find a maximal element $\f$. By our finiteness assumption, $\f$ is principal, so that $\f = \{\Phi \mid \Psi \impp \Phi\}$ for some $\Psi$. We claim that $\f \in \mathfrak{F}(\imp, \K^\dag)$, completing the proof. 

Let $\Phi \in \f$ and $\Phi \imp \Gamma$, then by \refax{DC-A}, $\Phi \impp \Gamma$, and so $\Gamma \in \f$; $\f$ is $\imp$-upwards-closed. Now let  $\{\Gamma,\Gamma'\} \subseteq \f$. Then we have that $\Psi \impp \Gamma$ and $\Psi \impp \Gamma'$ and $\Psi \notimpp \Lambda$, so by \refax{DC-B}, there exists some $\Psi'$ such that $\Psi' \impp \Psi$ and $\Psi' \imp \Gamma$ and $\Psi' \imp \Gamma'$ and $\Psi' \notimpp \Lambda$. Set $\f' = \{\Phi \mid \Psi' \impp \Phi\}$; we have $\f' \in F$ and $\f \subset \f'$, so by maximality $\f' = \f$. Thus, $\Psi' \in \f$; $\f$ is $\imp$-directed.
\end{tproof}

\begin{tproof}{thm:qStandad}
Towards necessity, assume quasi-standardness via $h$. Let $\Psi \imp \Lambda$. Since $(W,t)$ is a faithful, we have $t(\Psi) \subseteq t(\Lambda)$ and hence $h(t(\Psi)) \cap W^{\circ} \subseteq h(t(\Lambda)) \cap W^{\circ}$; it follows that $\Psi \impp \Lambda$, as required to show that \refax{DC-A} holds.

Now assume \refax{DC-A} holds.  Let $\W$ denote the set of all filters of $\imp$, with $t$ the inclusion map. As shown in the proof of Theorem \ref{thm:maxstatespace} this is a (maximal) faithful interpretation. Define $h$ as any monotone extension of the map
$$
h: t(\Psi) \mapsto \bigcup \{ t(\Lambda) \mid \Lambda \impp \Psi \}
$$
Note that over the image of $t$, $h$ is well defined---since if $t(\Psi) = t(\Psi')$, then $\Psi$ and $\Psi'$ are logically equivalent for $\imp$ and so, by \refax{DC-A}, for $\impp$ as well---and clearly monotone.

We will now show that $(\W^{\circ}, \Psi \mapsto h(t(\Psi))\cap \W^{\circ})$ is a faithful interpretation of $\ss$. Let $t^{\circ}$ denote the map $\Psi \mapsto h(t(\Psi))\cap \W^{\circ}$.

\begin{enumerate}[leftmargin=0cm]
  \item[\ref{faithful:subset}] First, let $\Phi \impp \Psi$. Let $w \in t^{\circ}(\Phi)$: there exists some $\Lambda$ such that $w \in t(\Lambda)$ and $\Lambda \impp \Phi$. Hence, $\Lambda \impp \Psi$, and so $w \in h(t(\Psi))$ and hence $t^{\circ}(\Psi)$. 

  Now let $\Phi \notimpp \Psi$. Then let $w = \{\Gamma \mid \Phi \imp \Gamma\}$; by definition $w \in t(\Phi)$ and so also $w \in h(t(\Phi))$. Now, take some arbitrary $\Lambda$ such that $\Lambda \impp \Psi$; it cannot be that $\Lambda \in w$, since then $\Phi \impp \Lambda \impp \Psi$, which we assumed was not the case (the first implication follows from \refax{DC-A}). By construction, for all such $\Lambda$, $w \notin t(\Lambda)$, and therefore, $w \notin h(t(\Psi))$. Further, notice that $w \notin h(t(\Psi))$ implies that $w \in W^{\circ}$, and so $w \in t^{\circ}(\Phi) - t^{\circ}(\Psi)$.

\item[\ref{faithful:null}] Let $\Phi$ be $\impp$-null and $w \in h(t(\Phi))$. Then there exists some $\Lambda$ such that $w \in t(\Lambda)$ and $\Lambda \impp \Phi$. It follows that $\Lambda$ is $\impp$-null as well. Then for any $\Psi$, we have $\Lambda \impp \Psi$, and so $w \in h(t(\Psi))$. Therefore $w \notin W^{\circ}$; $t^{\circ}(\Phi) = \emptyset$.

Now, let $\Phi$ be non $\impp$-null. Then there exists some $\Psi$ such that $\Phi \notimpp \Psi$, and therefore by \ref{faithful:subset} as just established, there exists some $w \in t^{\circ}(\Phi) - t^{\circ}(\Psi)$: $t^{\circ}(\Phi) \neq \emptyset$.

\item[\ref{faithful:disjoint}] Let $\w \in t^{\circ}(\Phi) \cap t^{\circ}(\Phi')$. Then there exists some $\Lambda, \Lambda'$ such that $\w \in t(\Lambda) \cap t(\Lambda')$ and such that $\Lambda \imp \Phi$ and $\Lambda' \imp \Phi'$. By the faithfulness of $(W,t)$, there exists some $\Gamma$ such that $\w \in t(\Gamma)$ and  $\Gamma \imp \Lambda$ and $\Gamma \imp \Lambda'$. By \refax{DC-A}, we have that $\Gamma \impp \Phi$ and $\Gamma \impp \Phi'$. Finally, since $\Gamma \impp \Gamma$, $w \in h(t(\Gamma))$ and hence $w \in t^{\circ}(\Gamma)$.
\qedhere
\end{enumerate}
\end{tproof}

\singlespace
\small
\bibliographystyle{aer}

\bibliography{FCT}



\end{document}